\documentclass{article}

\usepackage{arxiv}

\usepackage[utf8]{inputenc} 
\usepackage[T1]{fontenc}    
\usepackage{hyperref}       
\usepackage{url}            
\usepackage{booktabs}       
\usepackage{amsfonts}       
\usepackage{nicefrac}       
\usepackage{microtype}      
\usepackage{graphicx}
\usepackage{natbib}
\usepackage{doi}

\usepackage{amsmath}
\usepackage{bm, amsopn, amssymb, braket, enumitem, changepage, bookmark}
\usepackage[labelfont=bf]{caption}
\usepackage{threeparttable}
\usepackage[
singlelinecheck=false
]{caption}
\usepackage{multicol, array}
\usepackage[title]{appendix}
\usepackage[section]{placeins}
\usepackage[bottom]{footmisc}

\DeclareRobustCommand{\vect}[1]{\bm{#1}}
\title{Anomaly Detection in Aeronautics Data with Quantum-compatible Discrete Deep Generative Model}

\date{} 					

\author{Thomas Templin\hspace{.5mm}\textsuperscript{1}\href{https://orcid.org/0000-0001-5936-0009}{\includegraphics[scale=0.06]{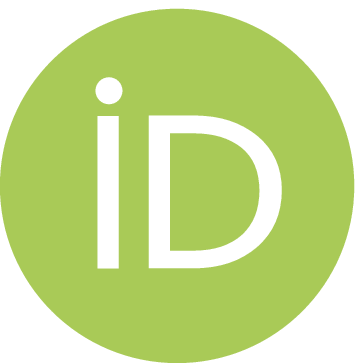}}, Milad Memarzadeh\hspace{.5mm}\textsuperscript{2}, Walter Vinci\hspace{.5mm}\textsuperscript{3}, P. Aaron Lott\hspace{.5mm}\textsuperscript{4},\\
\textbf{Ata Akbari Asanjan\hspace{.5mm}\textsuperscript{2}, Anthony Alexiades Armenakas\hspace{.5mm}\textsuperscript{4, 5}, and Eleanor Rieffel\hspace{.5mm}\textsuperscript{6}}\\
\\
\textsuperscript{1} Data Sciences Group, NASA Ames Research Center, Moffett Field, CA 94035, USA\\
\textsuperscript{2} Universities Space Research Association, Data Sciences Group,\\
NASA Ames Research Center, Moffett Field, CA 94035, USA\\
\textsuperscript{3} HP SCDS, 24009 Le\'{o}n, Spain\\
\textsuperscript{4} Universities Space Research Association, Quantum Artificial Intelligence Laboratory,\\
NASA Ames Research Center, Moffett Field, CA 94035, USA\\
\textsuperscript{5} Department of Physics, Harvard University, Cambridge, MA 02138, USA\\
\textsuperscript{6} Quantum Artificial Intelligence Laboratory, NASA Ames Research Center, Moffett Field, CA 94035, USA\\
\\
\textbf{Email:} \href{mailto: thomas.templin@nasa.gov}{thomas.templin@nasa.gov}
}
\renewcommand{\shorttitle}{Anomaly Detection with DVAE}

\hypersetup{
pdftitle={Anomaly Detection in Aeronautics Data with Quantum-compatible Discrete Deep Generative Model},
pdfsubject={Anomaly Detection with DVAE},
pdfauthor={Thomas Templin, Milad Memarzadeh, Walter Vinci, P. Aaron Lott, Ata Akbari Asanjan, Anthony Alexiades Armenakas, Eleanor Rieffel},
pdfkeywords={Generative modeling, Deep learning, Variational autoencoder, Anomaly detection, Restricted Boltzmann machine, Gibbs sampling, Quantum-assisted machine learning},
}

\begin{document}
\maketitle

\begin{abstract}
	Deep generative learning cannot only be used for generating new data with statistical characteristics derived from input data but also for anomaly detection, by separating nominal and anomalous instances based on their reconstruction quality. In this paper, we explore the performance of three unsupervised deep generative models---variational autoencoders (VAEs) with Gaussian, Bernoulli, and Boltzmann priors---in detecting anomalies in flight-operations data of commercial flights consisting of multivariate time series. We devised two VAE models with discrete latent variables (DVAEs), one with a factorized Bernoulli prior and one with a restricted Boltzmann machine (RBM) as prior, because of the demand for discrete-variable models in machine-learning applications and because the integration of quantum devices based on two-level quantum systems requires such models. The DVAE with RBM prior, using a relatively simple---and classically or quantum-mechanically enhanceable---sampling technique for the evolution of the RBM's negative phase, performed better than the Bernoulli DVAE and on par with the Gaussian model, which has a continuous latent space. Our studies demonstrate the competitiveness of a discrete deep generative model with its Gaussian counterpart on anomaly-detection tasks. Moreover, the DVAE model with RBM prior can be easily integrated with quantum sampling by outsourcing its generative process to measurements of quantum states obtained from a quantum annealer or gate-model device.
\end{abstract}


\newcolumntype{+}{>{\global\let\currentrowstyle\relax}}
\newcolumntype{^}{>{\currentrowstyle}}
\newcommand{\rowstyle}[1]{\gdef\currentrowstyle{#1}%
#1\ignorespaces
}

\interfootnotelinepenalty=10000

\section{Introduction} \label{introduction}

The field of machine learning has experienced an explosion in the development of deep-learning methods at the beginning of the 21st century, due to the flexibility, scalability, and superior performance of deep learning in classification, prediction, data generation, anomaly detection, and other applications \citep{hinton2006fast, bengio2006greedy, lecun2015deep, goodfellow2016deep}. The phenomenal success of deep learning, which refers to machine-learning techniques that use artificial-neural-network models with many layers, has been enabled by the widespread availability of specialized graphics processing units (GPUs) to perform computing-intensive linear-algebra operations on vectors, matrices, and tensors. These algebraic structures hold the numerical values of the input, intermediate, and output layers of the (deep) network, as well as the values of the biases and weights that are applied to the hidden and output units (neurons) of the network. During network training, manipulations of vectors, matrices, and higher-order tensors are performed ubiquitously to compute the network's output in the forward pass and the gradients to update the biases and weights in the backward pass \citep{lecun2015deep, witten2017practical}.

One way to characterize the training process of a neural network is to differentiate between supervised and unsupervised learning \citep{dayan2005theoretical, witten2017practical}. In supervised learning, a `teacher' imposes a set of desired input-output relationships on the network. For example, the training set might contain an extra column that specifies the desired output of the network, such as a class label. The class label or other supervisory information is not available during testing, when the performance of the network is evaluated. In unsupervised learning, no such oversight is provided, and the network's response is self-organized and solely relies on the interplay between external input, intrinsic connectivity, network dynamics, and the value of a cost function that the network attempts to minimize. Unsupervised learning is computationally more complex than supervised learning and still a largely unresolved problem in machine learning. It has attracted considerable research effort \citep{hinton1995wake, bengio2006greedy, vincent2008extracting} because it holds the potential to uncover the statistical structure and hidden correlations of large unlabeled datasets, which constitute the predominant form of today's data.

Generative modeling is a machine-learning technique widely used in unsupervised learning. Generative modeling attempts to estimate the probability distribution of a dataset. To accomplish this goal, generative models frequently employ a set of latent variables that represent unobserved factors that influence the values of observed variables. Deep generative models such as generative adversarial networks \citep{goodfellow2020generative}, variational autoencoders (VAEs) \citep{kingma2013auto}, and deep belief networks \citep{hinton2006reducing} have been widely applied to machine-learning use cases in science and engineering. In the studies reported in this paper, we use VAEs for generative modeling because a VAE possesses an efficient inference mechanism, incorporates regularization via a prior, maximizes a lower bound on the log likelihood, and allows estimation of the log likelihood via importance sampling \citep{kingma2013auto, burda2015importance}. VAEs employ the evidence lower bound (ELBO) as a variational lower bound on the exact log likelihood. The (negative) ELBO is a well-defined, fully differentiable, loss function whose gradients are used to efficiently optimize network weights through backpropagation, permitting competitive performance in mining large datasets. 

The majority of VAE and other generative-model designs reported in the literature use continuous latent spaces because of the widespread applicability of the normal distribution, which is continuous, due to the central limit theorem and the difficulty of propagating gradients through discrete variables. However, many deep-learning use cases rely on discrete latent variables to represent the required distributions, such as in applications in supervised and unsupervised learning, attention models, language models, and reinforcement learning \citep{kingma2014semi, jang2016categorical, maaloe2017semi}. In particular, if the values of latent variables are to be computed by quantum computers, the latent variables need to be discrete because projective qubit measurements in the computational basis produce eigenvalues of -1 or +1. See supplementary section \hyperref[suppl:1]{S1} for a more in-depth account of the importance of discrete-variable models.

In previous studies, discrete VAEs (DVAEs) and quantum VAEs were used to generate new data from samples from the VAE's latent space after the VAE had been trained on a dataset such as MNIST or Omniglot, and the quality of generation (fit of the VAE's model distribution to the distribution of the input data) was assessed by estimating the log likelihood of test data \citep{rolfe2016discrete, vahdat2018dvae++, vahdat2018dvae, khoshaman2019quantum, khoshaman2018gumbolt, vinci2020path, vahdat2020undirected}. In the studies reported in this paper, we use VAE models with continuous and discrete latent space to detect anomalies in aeronautics data. The datasets used comprise primarily 1-Hz recordings of operationally significant flight metrics from commercial flights. Subject matter experts analyzed the recorded data and identified operationally and safety-relevant anomalies during takeoff and approach to landing. The identification of flight-operations anomalies is important because they can foreshadow potentially serious aviation incidents or accidents.

The application of VAEs to anomaly-detection tasks has become increasingly popular in recent years. \citet{an2015variational} suggested an anomaly-detection method in which the anomaly score of a VAE is used as a Monte Carlo estimate of the reconstruction log likelihood (called “reconstruction probability” in the paper). \citet{xu2018unsupervised} used a VAE for the detection of anomalies in univariate time series, preprocessed with sliding time windows, representing seasonal key performance indicators in web applications. Based on the success of deep recurrent neural networks (RNNs) in machine-learning applications with sequential data, several studies have incorporated RNNs in VAEs by equipping the VAE's encoder and decoder with long short-term memory constructs \citep{chen2019sequential, Wang2020advae, Zhang2019time, Zhang2019deep, park2018multimodal}. The LSTM-VAE approach was also applied to anomaly detection in telemetry data from the Soil Moisture Active Passive (SMAP) satellite and the Mars Curiosity rover \citep{su2019robust}. However, the training of a VAE equipped with an RNN architecture on multidimensional time series is computationally costly and may overlook local temporal dependencies. To remedy these shortcomings, \citet{memarzadeh2020unsupervised} designed a convolutional VAE (CVAE) and tested its performance in detecting anomalies in time series of various Yahoo! benchmark datasets and in time series spanning the takeoff phase of commercial flights, a task on which the model achieved state-of-the-art performance.

We developed convolutional VAEs with Gaussian, Bernoulli, and Boltzmann priors. The VAE with Gaussian prior has a continuous latent space, whereas the models with Bernoulli or Boltzmann prior have a discrete latent space. The Boltzmann prior is implemented as a restricted Boltzmann machine (RBM) \citep{smolensky1986information}, that is, as a network of stochastic binary units with full connectivity between visible and hidden units but no connectivity between visible units or between hidden units. The VAE with Gaussian prior and the RBM network of the VAE model with RBM prior are derivations of the CVAE model presented in \citet{memarzadeh2020unsupervised} and of the DVAE model depicted in \citet{vinci2020path}, respectively. Overall, our studies aim to determine and compare the anomaly-detection performance of the Gaussian, Bernoulli, and RBM models.\footnote[1]{For simplicity’s sake, we frequently refer to the VAE models with Gaussian, Bernoulli, and RBM priors as the Gaussian, Bernoulli, and RBM models, respectively, in this paper; the longer, more correct, expression is used interchangeably with the abbreviated version.} We want to find out if the anomaly-detection performance of a VAE with discrete latent space is competitive with that of a VAE with Gaussian prior and continuous latent variables, the standard choice of VAE type. Also, if a classical deep generative model with discrete latent variables exhibits a performance that is comparable or superior to that of a continuous-variable counterpart, it is worth exploring if a quantum-enhanced version of the discrete model can achieve a performance that exceeds that of the fully classical discrete model.

We report the results of three sets of experiments. Using a dataset with a drop-in-airspeed anomaly during takeoff, our baseline study explores the behavior of the VAE models during training, with an emphasis on the training behavior of the RBM model, and compares the models' anomaly-detection performance when operating with either optimized or nonoptimal hyperparameters, as given by the performance metrics precision, recall, and F1 score. Our second study investigates the ability of our trained models to generalize (transfer) to a new dataset containing the same anomaly without re-tuning of hyperparameters and without or with post-training on the new dataset. Finally, we examine if the performance of the RBM model is robust to changes in anomaly type and phase of flight, by evaluating the model's performance on a new dataset with delay-in-flap-deployment anomaly during approach to landing; for this study, the model's hyperparameters were re-tuned and the model was re-trained on the dataset used.

The structure of the paper is as follows. In section \ref{causal}, we review causal generative models, that is, probabilistic models that reconstruct input data from latent variables. The concept of variational inference, as an approximation to an intractable posterior distribution, and prior distributions used in generative modeling with continuous and discrete latent variables are described. Section \ref{vaes} covers VAEs with continuous and discrete latent spaces. We describe the $\beta$-VAE, used in our experiments, a type of VAE model that allows a weighting of the autoencoding and Kullback-Leibler (KL)-divergence terms in the variational ELBO objective. Alternative formulations of RBM prior networks in the VAE's latent space are also introduced. In section \ref{evaluation}, we describe the methodology we used to evaluate our models' anomaly-detection performance. In section \ref{results}, we present the experimental findings of our three studies, outlined in the preceding paragraph. We discuss model design and performance in section \ref{discussion} and present our conclusions in section \ref{conclusions}. The appendix and supplementary material contain additional information on concepts and experiments.

\section{Causal generative modeling} \label{causal}

\begin{figure}[h]
    \captionsetup{singlelinecheck = true, justification=justified, font=footnotesize, labelsep=period, width=1\textwidth}
    \centering
    \includegraphics[width=0.85\textwidth]{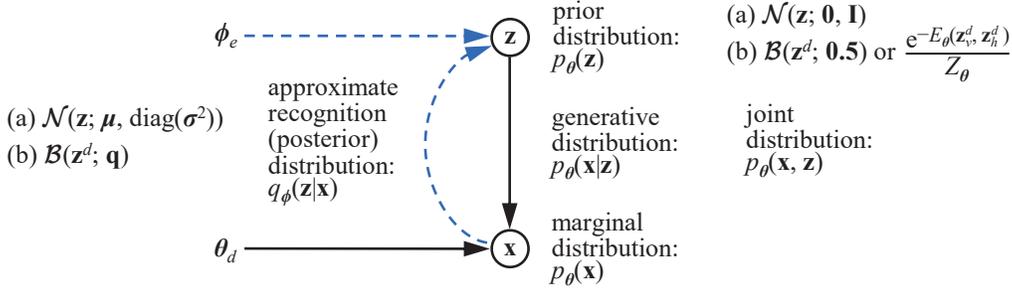}
    \caption{Generative models with latent variables, $\mathbf{z}$, can be represented as probabilistic graphical models that depict conditional relationships among variables. In a directed generative model, the model distribution, $p_{\bm{\theta}}(\mathbf{x}, \mathbf{z})$ $= p_{\bm{\theta}}(\mathbf{x}|\mathbf{z}) p_{\bm{\theta}}(\mathbf{z})$, is explicitly factored into the generative (decoder) distribution, $p_{\bm{\theta}}(\mathbf{x}|\mathbf{z})$, and the model's prior distribution over latent variables, $p_{\bm{\theta}}(\mathbf{z})$. The marginal distribution, $p_{\bm{\theta}}(\mathbf{x})$, is obtained by marginalizing (integrating or summing) over the latent variables. Since the computation of the true posterior, $p_{\bm{\theta}}(\mathbf{z}|\mathbf{x})$, is intractable, variational inference substitutes it with an approximating posterior (encoder), $q_{\bm{\phi}}(\mathbf{z}|\mathbf{x})$. Black, solid arrows denote the generative model and blue, dashed arrows the variational approximation to the intractable posterior. In the VAE models used to conduct the experiments described in this paper, the posterior and prior over latent variables are either (a) continuous Gaussian distributions or (b) discrete Bernoulli or Boltzmann distributions. We use the symbolic expression $z^d$ to denote a \textit{discrete} latent variable.}
    \addtocounter{figure}{-1}
    \phantomcaption
    \label{fig:1}
\end{figure}

The goal of generative modeling is to estimate the probability distribution of the input data, $p(\mathbf{x})$, which is unknown but assumed to exist. The distribution of synthetic data points, estimated by the model, is called the marginal distribution, $p_{\bm{\theta}}(\mathbf{x})$ (where $\bm{\theta}$ denotes the model parameters). The goal is to make $p_{\bm{\theta}}(\mathbf{x})$ as close to $p(\mathbf{x})$ as possible. In order to accomplish this objective, representational modeling computationally analyzes the statistical structure of a dataset and attempts to identify a set of latent (unobserved) variables $\mathbf{z}$ that represent the dominant features of the dataset. Latent variables are also known as `causes' \citep{dayan2005theoretical}. A graphical model of a directed generative model with latent variables is depicted in figure \ref{fig:1}. The generative model (decoder) reconstructs the input data from the latent variables \citep{dayan2005theoretical, kingma2019introduction}. The statistics of the generative model's output are given by the marginal distribution:

\begin{equation} \label{eq:1}
    p_{\bm{\theta}}(\mathbf{x}) = \int_{\mathbf{z}} p_{\bm{\theta}}(\mathbf{x}, \mathbf{z}) \mathrm{d}\mathbf{z} = \int_{\mathbf{z}} p_{\bm{\theta}}(\mathbf{x}|\mathbf{z}) p_{\bm{\theta}}(\mathbf{z}) \mathrm{d}\mathbf{z}.
\end{equation}

\noindent
Here, the joint distribution of the input variables $\mathbf{x}$ and the latent variables $\mathbf{z}$, $p_{\bm{\theta}}(\mathbf{x}, \mathbf{z})$, defines the model distribution. In a directed generative model, the model distribution is explicitly factored into the generative distribution $p_{\bm{\theta}}(\mathbf{x}|\mathbf{z})$, the distribution of the generative model's output given a latent-space realization, and the model's prior distribution $p_{\bm{\theta}}(\mathbf{z})$. The marginal distribution is then obtained by integrating (`marginalizing') over the latent variables (see figure \ref{fig:1}). If the latent variables are discrete, the integration in (\ref{eq:1}) is replaced by summation. Generative modeling is well suited for unsupervised learning: Lacking supervisory information, a model's performance is determined by the ability of its latent variables to represent and reproduce the statistical structure of the input variables \citep{dayan2005theoretical}.

In representational learning, input-dependent latent variables (the `code') are identified by a second model, the recognition model (encoder). The lower the number of latent variables is, the more compressed is the model's latent space, with the degree of compression given by the ratio of input to latent variables. The statistical distribution of the recognition model's output is called the recognition or posterior distribution, $p_{\bm{\theta}}(\mathbf{z}|\mathbf{x})$, the probability of the set of latent variables conditioned on the input data (see figure \ref{fig:1}). The posterior distribution is given by Bayes' rule as \citep{kingma2019introduction}:

\begin{equation} \label{eq:2}
    p_{\bm{\theta}}(\mathbf{z}|\mathbf{x}) = \frac{p_{\bm{\theta}}(\mathbf{x}, \mathbf{z})}{p_{\bm{\theta}}(\mathbf{x})} = \frac{p_{\bm{\theta}}(\mathbf{x}|\mathbf{z}) p_{\bm{\theta}}(\mathbf{z})}{p_{\bm{\theta}}(\mathbf{x})}.
\end{equation}

\subsection{Variational inference} \label{variational}

Models that allow the tractable computation of the posterior distribution are called invertible and those that do not noninvertible \citep{dayan2005theoretical}. Noninvertible models do not allow gradient computations and concomitant optimization. Deep latent-variable models are noninvertible because no analytic solution or efficient estimation procedure exists for the marginal probability given in (\ref{eq:1}). Since the marginal is intractable, the posterior given in (\ref{eq:2}) is as well as it requires the marginal in its denominator \citep{kingma2019introduction}. Approximate-inference techniques approximate the true posterior $p_{\bm{\theta}}(\mathbf{z}|\mathbf{x})$ and marginal $p_{\bm{\theta}}(\mathbf{x})$ \citep{dayan2005theoretical, kingma2019introduction}. The approximate posterior is written as $q_{\bm{\phi}}(\mathbf{z}|\mathbf{x})$ (see figure \ref{fig:1}). Variational inference is such an approximation technique. In variational inference, the marginal log likelihood is decomposed as follows \citep{kingma2013auto, kingma2019introduction}:

\begin{equation} \label{eq:3}
    \mathrm{log} \; p_{\bm{\theta}}(\mathbf{x}) = D_{\mathrm{KL}}(q_{\bm{\phi}}(\mathbf{z}|\mathbf{x})||p_{\bm{\theta}}(\mathbf{z}|\mathbf{x})) + \mathcal{L}(\bm{\theta}, \bm{\phi};\mathbf{x}).
\end{equation}

\noindent
The expression $D_{\mathrm{KL}}(p||q) \equiv \mathbb{E}_{p} \, \mathrm{log}[p/q]$ stands for the Kullback-Leibler (KL) divergence, an asymmetric `distance' between two probability distributions. The term $\mathcal{L}(\bm{\theta}, \bm{\phi};\mathbf{x})$ denotes the ELBO---a variational approximation, from below, to the true marginal log likelihood. As the KL divergence is non-negative, the ELBO is the greater and closer to the log likelihood, i.e., the bound tighter, the closer the approximating posterior is to the true posterior. Hence, maximizing the ELBO simultaneously increases the (log) likelihood and decreases the distance between the variational and true posteriors.

\subsection{Prior distributions} \label{priors}

The simplest generative-model priors are factorized standard normal or Bernoulli distributions:

\begin{equation} \label{eq:5}
    \begin{split}
        p_{\bm{\theta}}(\mathbf{z}) &= \mathcal{N}(\mathbf{z}; \mathbf{0},\mathbf{I}) \equiv \prod_{l=1}^{L} \mathcal{N}(z_l; 0,1),\cr
        p_{\bm{\theta}}(\mathbf{z}^d) &= \mathcal{B}(\mathbf{z}^d; \mathbf{0.5}) \equiv \prod_{l=1}^{L} \mathcal{B}(z_l^d; 0.5),
    \end{split}
\end{equation}

\noindent
where the subscript $l$ denotes a latent variable. Also, we use the symbolic expression $z^d$ to denote a discrete latent variable (to delineate it from a continuous latent variable, expressed as $z$). The posterior distributions factorize accordingly \citep{khoshaman2019quantum}.

Boltzmann machine (BM) priors are more expressive and capable of representing complex multi-model probability distributions \citep{khoshaman2019quantum}. Since Boltzmann priors are intractable in models with numbers of variables useful for practical problems, it is common to use Markov chain Monte Carlo (MCMC) sampling to estimate gradients. The efficiency of MCMC sampling is greatly increased \citep{khoshaman2018gumbolt} by stipulating a bipartite connectivity between the groups of visible units $\mathbf{z}_v^d$ and hidden units $\mathbf{z}_h^d$ of the BM, without lateral connections between the units within either group, i.e., an RBM \citep{smolensky1986information}. The visible and hidden units correspond to the input and latent variables, respectively, of an undirected generative model and are binary (0 or 1) in the RBMs we employed in the experiments reported in this paper. An RBM prior is given by \citep{welling2004exponential, salakhutdinov2007restricted, khoshaman2019quantum}:

\begin{equation}  \label{eq:6}
    \begin{split}
        p_{\bm{\theta}}(\mathbf{z}_v^d, \mathbf{z}_h^d) &= \mathrm{e}^{-E_{\bm{\theta}}(\mathbf{z}_v^d, \mathbf{z}_h^d)} / Z_{\bm{\theta}}, \quad Z_{\bm{\theta}} \equiv \sum_{\mathbf{z}_v^d, \mathbf{z}_h^d} \mathrm{e}^{-E_{\bm{\theta}}(\mathbf{z}_v^d, \mathbf{z}_h^d)},\cr
        E_{\bm{\theta}}(\mathbf{z}_v^d, \mathbf{z}_h^d) &= -(\mathbf{z}_v^d)^\mathrm{T}\mathbf{Wz}_h^d - \mathbf{a}^\mathrm{T} \mathbf{z}_v^d - \mathbf{b}^\mathrm{T} \mathbf{z}_h^d.
    \end{split}
\end{equation}

\noindent
In (\ref{eq:6}), $E_{\bf{\theta}}(\mathbf{z}_v^d, \mathbf{z}_h^d)$ is the energy function of visible and hidden units and $Z_{\bm{\theta}}$ the normalizing constant (partition function) of prior $p_{\bm{\theta}}(\mathbf{z}_v^d, \mathbf{z}_h^d)$; $\mathbf{W}$, $\mathbf{a}$, and $\mathbf{b}$ are the weight matrix between visible and hidden units and the visible and hidden bias vectors, respectively. The conditional distributions of the hidden given the visible units and of the visible given the hidden units are then given by \citep{witten2017practical}:

\begin{equation}  \label{eq:7}
    \begin{split}
        p_{\bm{\theta}}(\mathbf{z}_h^d|\mathbf{z}_v^d) &= \prod_{l=1}^{L} p_{\bm{\theta}}((z_h^d)_l|\mathbf{z}_v^d) = \prod_{l=1}^{L} \sigma(b_l + \mathbf{W}_{\cdot l}^{\mathrm{T}} \, \mathbf{z}_v^d),\cr
        p_{\bm{\theta}}(\mathbf{z}_v^d|\mathbf{z}_h^d) &= \prod_{k=1}^{K} p_{\bm{\theta}}((z_v^d)_k|\mathbf{z}_h^d) = \prod_{k=1}^{K} \sigma(a_k + \mathbf{W}_{k \cdot} \, \mathbf{z}_h^d),
    \end{split}
\end{equation}

\noindent
where $\sigma$ denotes the logistic function and $\mathbf{W}_{\cdot l}^{\mathrm{T}}$ is a vector consisting of the transpose of the \textit{l}th column of weight matrix \textbf{W} and $\mathbf{W}_{k \cdot}$ a vector consisting of the \textit{k}th row of \textbf{W}. Because of the absence of lateral connections between visible units and between hidden units, the conditional probabilities in (\ref{eq:7}) can be determined in one fell swoop, using block Gibbs sampling. In block Gibbs sampling, the values of all hidden units are updated at once by sampling from their conditional distribution given the visible units [top equation of (\ref{eq:7})]. Then, the values of the visible units are updated analogously [bottom equation of (\ref{eq:7})]. This process can be repeated for an arbitrary number of iterations. When Gibbs sampling is performed for an infinite number of steps, it is guaranteed to converge to the stationary distribution $p_{\bm{\theta}}(\mathbf{z}_v^d, \mathbf{z}_h^d)$ of the RBM model \citep{hinton2012practical, fischer2014training}, and computationally efficient techniques to learn the model distribution have been developed \citep{hinton2002training, tieleman2008training, tieleman2009using}.

\section{Variational autoencoders} \label{vaes}

VAEs are directed generative models with latent variables that approximate the intractable true posterior $p_{\bm{\theta}}(\mathbf{z}|\mathbf{x})$ via variational inference and maximize an ELBO objective $\mathcal{L}(\bm{\theta}, \bm{\phi};\mathbf{x})$ (see section \ref{variational} and figure \ref{fig:1}). The ELBO can be re-written as \citep{kingma2013auto}:

\begin{equation} \label{eq:4}
    \mathcal{L}(\bm{\theta}, \bm{\phi};\mathbf{x}) = \mathbb{E}_{\mathbf{z} \sim q_{\bm{\phi}}(\mathbf{z}|\mathbf{x})}[\mathrm{log} \; p_{\bm{\theta}}(\mathbf{x}|\mathbf{z})] - D_{\mathrm{KL}}(q_{\bm{\phi}}(\mathbf{z}|\mathbf{x})||p_{\bm{\theta}}(\mathbf{z})).
\end{equation}

\noindent
The first term of (\ref{eq:4}) is the autoencoding term. Maximizing it maximizes the fidelity of reconstruction because the greater the autoencoding term is, the greater is the similarity between the data distribution $p(\mathbf{x}$) and the generative distribution $p_{\bm{\theta}}(\mathbf{x}|\mathbf{z})$ when $\mathbf{z}$ is sampled from the approximate posterior distribution $q_{\bm{\phi}}(\mathbf{z}|\mathbf{x})$ of the encoder. Conversely, the second term is maximized by minimizing the KL divergence between the approximate posterior and prior, which corresponds to minimizing the mutual information between $\mathbf{x}$ and $\mathbf{z}$. Consequently, the autoencoding term attempts to maximize the mutual information between data and latents and the KL term seeks to minimize it. Eventually, the latent-space's information content will depend on the trade-off between the two terms, which, in turn, is determined by the flexibility and expressiveness of the variational approximation $q_{\bm{\phi}}(\mathbf{z}|\mathbf{x})$, the structure of the generative model, and the training method \citep{khoshaman2019quantum, vinci2020path}.

Given training set $\mathcal{D} = \{\mathbf{x}^{(n)}\}_{n=1}^N$ consisting of \textit{N} i.i.d. samples from $p(\mathbf{x})$, the ELBO for a minibatch of training data is given as the average of the ELBOs of the minibatch instances \citep{kingma2013auto}:

\begin{equation} \label{eq:4a}
    \mathcal{L}(\bm{\theta}, \bm{\phi};\mathcal{M}^{(i)}) = \frac{1}{M} \sum_{\mathbf{x} \in \mathcal{M}^{(i)}} \mathcal{L}(\bm{\theta}, \bm{\phi};\mathbf{x}),
\end{equation}

\noindent
where the minibatch $\mathcal{M}^{(i)} = \{\mathbf{x}^{(i, m)}\}_{m=1}^M$ contains \textit{M} data points randomly sampled from $\mathcal{D}$ with \textit{N} data points.

\subsection{VAE with factorized Gaussian prior} \label{gaussian}

The ELBO objective given in (\ref{eq:4}) contains expectations of functions of the latent variables $\mathbf{z}$ with regard to the variational posterior $q_\phi(\mathbf{z}|\mathbf{x})$, which can be written as $\mathbb{E}_{\mathbf{z} \sim q_{\bm{\phi}}(\mathbf{z}|\mathbf{x})}[f(\mathbf{z})]$, where $f$ denotes an arbitrary function. To train the model with minibatch stochastic gradient descent starting from random initializations of the model parameters $\bm{\theta}$ and $\bm{\phi}$, we need to calculate gradients of such expectations \citep{kingma2019introduction, khoshaman2019quantum}. Procedures to obtain unbiased gradients with respect to $\bm{\theta}$ and $\bm{\phi}$ are described in Appendix \hyperref[app:a]{A}.

We estimated the Gaussian parameters $\bm{\mu}$ and $\mathrm{log} \; \bm{\sigma}^2$ by means of linear layers at the top of the VAE's encoder (see Appendix \hyperref[app:d]{D}). The $\mathrm{log} \; \bm{\sigma}^2$ estimate was then routed through a softplus activation. Hence, the mean and log variance of the approximate posterior, $\bm{\mu}$ and $\mathrm{log} \; \bm{\sigma}^2$, are nonlinear functions of the input data \textbf{x} and the variational parameters $\bm{\phi}$ \citep{kingma2013auto}.

Moreover, when the VAE prior is given by a factorized Gaussian, the KL-divergence term in the ELBO objective [(\ref{eq:4})] can be expressed in closed form. The ELBO is then estimated as:

\begin{equation} \label{eq:13}
    \mathcal{L}(\bm{\theta}, \bm{\phi};\mathbf{x}) \simeq \frac{1}{S} \sum_{s=1}^{S} \mathrm{log} \; p_{\bm{\theta}}(\mathbf{x}|\mathbf{z}^{(s)}) + \frac{1}{2} \sum_{l=1}^{L} (1 + \mathrm{log} \: \sigma_l^2 - \mu_l^2 - \sigma_l^2),
\end{equation}

\noindent
where $\mathbf{z}^{(s)} = \bm{\mu} + \bm{\sigma} \odot \bm{\epsilon}^{(s)}$, $\bm{\epsilon}^{(s)} \sim \mathcal{N}(\mathbf{0}, \mathbf{I})$, and $l$ indexes a latent variable \citep{kingma2013auto}.

\subsection{$\vect{\beta}$-VAE} \label{beta}

\citet{higgins2017beta} modified the VAE objective to reduce the entanglement between latent variables. Each latent variable $z_l$ is to represent a meaningful domain-specific attribute that varies along a continuum when $z_l$ is varied. In order to promote this disentangling property in the latent variables $\mathbf{z} \sim q_{\bm{\phi}}(\mathbf{z}|\mathbf{x})$, the authors introduce a constraint over the posterior $q_{\bm{\phi}}(\mathbf{z}|\mathbf{x})$ from which they are derived, making it more similar to a prior $p_{\bm{\theta}}(\mathbf{z})$, and thus restrain latent-space capacity and stimulate statistical independence between individual latent-space variables. The ELBO objective [(\ref{eq:4})] of a $\beta$-VAE is given as (see Appendix \hyperref[app:b]{B} for derivation):

\begin{equation} \label{eq:16}
    \mathcal{L}(\bm{\theta}, \bm{\phi};\mathbf{x}, \beta) = \mathbb{E}_{\mathbf{z} \sim q_{\bm{\phi}}(\mathbf{z}|\mathbf{x})}[\mathrm{log} \; p_{\bm{\theta}}(\mathbf{x}|\mathbf{z})] - \beta \, D_{\mathrm{KL}}(q_{\bm{\phi}}(\mathbf{z}|\mathbf{x})||p_{\bm{\theta}}(\mathbf{z})).
\end{equation}

\noindent
The parameter $\beta$ is used to balance the trade-off between the fit of the reconstructed data, $\hat{\mathbf{x}}$, to the input data, $\mathbf{x}$, imposed by the autoencoding (reconstruction) term (low $\beta$), and the fit of the posterior, $q_{\bm{\phi}}(\mathbf{z}|\mathbf{x})$, to the prior, $p_{\bm{\theta}}(\mathbf{z})$, via the KL term (high $\beta$). If $\beta = 0$, the $\beta$-VAE model is identical to an autoencoder, and if $\beta = 1$, the model corresponds to a regular VAE. We would like to note that we do not use $\beta$ to explicitly disentangle the latent space but employ it as a regularization hyperparameter that requires tuning, an approach pioneered in \citet{memarzadeh2020unsupervised}.

\subsection{VAE with discrete latent space} \label{discrete}

Several approaches have been developed to circumvent the non-differentiability problem affecting models with discrete latent units \citep{mnih2014neural, paisley2012variational, gu2015muprop, bengio2013estimating}. In VAE models, the reparameterization trick has been extended by either the incorporation of smoothing functions \citep{rolfe2016discrete} or the relaxation of discrete latent variables into continuous ones \citep{jang2016categorical, maddison2016concrete, khoshaman2018gumbolt}. In this work, we employ the Gumbel-softmax trick, which relaxes a discrete categorical distribution into a continuous concrete (or Gumbel-softmax) distribution \citep{maddison2016concrete, jang2016categorical}.

If each individual latent variable is sampled as $z^d \sim \mathcal{B}(p)$, where $\mathcal{B}$ denotes the Bernoulli distribution and $p$ its parameter, the concrete relaxation can be expressed as \citep{maddison2016concrete}:

\begin{equation} \label{eq:17}
    \mathbf{z} = \sigma((\mathrm{log} \; \bm{\alpha} + \mathrm{log} \; \bm{\rho} - \mathrm{log} (1 - \bm{\rho}))/\lambda),
\end{equation}

\noindent
where $\sigma$ denotes the logistic function, $\sigma(x) = 1/(1 + e^{-x})$, $\alpha$ the odds of the Bernoulli probability, $\alpha = p/(1-p)$, $\rho$ a continuous uniform random variable on the interval [0, 1], and $\lambda$ the temperature of the concrete distribution. The temperature parameter $\lambda$ controls the degree of continuous relaxation: the greater $\lambda$ is, the greater is the relaxation and departure from a Bernoulli random variable, whereas small $\lambda$ values produce continuous variates close to 0 and 1. In our studies, we consistently use a $\lambda$ of 0.1, which we determined as a hyperparameter with good performance and which introduces only a small bias. The Gumbel-softmax relaxation was only applied during training and not during evaluation (validation or testing) because differentiability of the objective function is only required during the training phase and we sought to retain, where possible, the discrete character of the model and avoid the (slight) estimation bias introduced by the continuous relaxation. We estimated the log odds of the approximate relaxed Bernoulli posterior probabilities, $\mathrm{log} \; \bm{\alpha}^{\mathbf{q}}$, by means of a linear layer at the top of the VAE's encoder (see Appendix \hyperref[app:d]{D}).

Discrete VAEs can be implemented with Bernoulli [bottom equation in (\ref{eq:5})] or RBM [(\ref{eq:6})] priors. Based on the probability mass function of a Bernoulli random variable, $\mathcal{B}(k; p) = p^k \; (1-p)^{1-k}$, the KL term of the ELBO of a VAE with a Bernoulli prior can be expressed as:

\begin{equation} \label{eq:18}
    \begin{split}
        &D_{\mathrm{KL}}(q_{\bm{\phi}}(\mathbf{z}^d|\mathbf{x})||p_{\bm{\theta}}(\mathbf{z}^d)) = \mathbb{E}_{\mathbf{z}^d \sim q_{\bm{\phi}}(\mathbf{z}^d|\mathbf{x})}[\mathrm{log} \frac{q_{\bm{\phi}}(\mathbf{z}^d|\mathbf{x})}{p_{\bm{\theta}}(\mathbf{z}^d)}]\cr
        &= \mathbb{E}_{\mathbf{z}^d \sim \mathcal{B}(\mathbf{q})}[\sum_{l=1}^{L} \mathrm{log} \frac{q_l^{z_l^d} (1 - q_l)^{1 - z_l^d}}{0.5^{z_l^d} (1 - 0.5)^{1 - z_l^d}}]\cr
        &= \mathbb{E}_{\mathbf{z}^d \sim \mathcal{B}(\mathbf{q})}[\sum_{l=1}^{L} (z_l^d \; \mathrm{log} \; q_l + (1 - z_l^d) \, \mathrm{log}(1 - q_l) - z_l^d \; \mathrm{log} \; 0.5 - (1 - z_l^d) \, \mathrm{log}(1 - 0.5))]\cr
        &= \mathbb{E}_{\mathbf{z}^d \sim \mathcal{B}(\mathbf{q})}[\sum_{l=1}^{L} (-z_l^d \; \mathrm{log} \frac{0.5}{q_l} - (1 - z_l^d) \, \mathrm{log} \frac{1 - 0.5}{1 - q_l})],
    \end{split}
\end{equation}

\noindent
where $q_l$ stands for the parameter of the latent Bernoulli variable $z_l^d \sim \mathcal{B}(q_l)$, considered to be \textit{discrete}, and 0.5 is the parameter of the Bernoulli prior distribution, $\mathcal{B}(p=0.5)$. The KL term can be implemented using the \texttt{BCEWithLogitsLoss($\cdot$)} function of the PyTorch \texttt{nn} module with the log-odds parameter vector of the respective \textit{relaxed} Bernoulli distribution [$\mathrm{log} \; \bm{\alpha}^{\mathbf{q}}$ (approximate posterior) or $\mathrm{log} \; \bm{\alpha}^{\mathbf{p}} = \mathbf{0}$ (prior)] and the \textit{relaxed} latent vector \textbf{z} (training) or \textit{hard} latent vector $\mathbf{z}^d$ (evaluation) as input arguments. We used this (stochastic) method [Monte Carlo estimator of (\ref{eq:18})] in all experiments in which the VAE had a Bernoulli prior. An analytic expression for the KL term in the ELBO objective [(\ref{eq:4})] of a VAE with Bernoulli prior is given in Appendix \hyperref[app:c]{C}.

Since the applied concrete (Gumbel-softmax) relaxation replaces discrete latent variables $\mathbf{z}^d$ with continuous variables \textbf{z}, the KL term of the ELBO of a VAE with a relaxed Bernoulli posterior and an RBM prior can be expressed as:

\begin{equation} \label{eq:21}
    \begin{split}
        &D_{\mathrm{KL}}(q_{\bm{\phi}}(\mathbf{z}|\mathbf{x})||p_{\bm{\theta}}(\mathbf{z}))\cr
        &= \mathbb{E}_{\mathbf{z} \sim q_{\bm{\phi}}(\mathbf{z}|\mathbf{x})}[\mathrm{log} \frac{q_{\bm{\phi}}(\mathbf{z}|\mathbf{x})}{p_{\bm{\theta}}(\mathbf{z})}]\cr
        &= \mathbb{E}_{\mathbf{z} \sim q_{\bm{\phi}}(\mathbf{z}|\mathbf{x})}[\mathrm{log} \; q_{\bm{\phi}}(\mathbf{z}|\mathbf{x})] - \mathbb{E}_{\mathbf{z} \sim q_{\bm{\phi}}(\mathbf{z}|\mathbf{x})}[\mathrm{log} \; p_{\bm{\theta}}(\mathbf{z})]\cr
        &= \mathbb{E}_{\mathbf{z} \sim q_{\bm{\phi}}(\mathbf{z}|\mathbf{x})}[\mathrm{log} \; q_{\bm{\phi}}(\mathbf{z}|\mathbf{x})] - \mathbb{E}_{\mathbf{z} \sim q_{\bm{\phi}}(\mathbf{z}|\mathbf{x})}[\mathrm{log} \frac{\mathrm{e}^{-E_{\bm{\theta}}(\mathbf{z})}}{Z_{\bm{\theta}}}]\cr
        &= \mathbb{E}_{\bm{\rho} \sim U(0, 1)}[\mathrm{log} \; q_{\bm{\phi}}(\mathbf{z(\bm{\phi}, \bm{\rho}})|\mathbf{x})] - \mathbb{E}_{\bm{\rho} \sim U(0, 1)}[\mathrm{log} \frac{\mathrm{e}^{-E_{\bm{\theta}}(\mathbf{z(\bm{\phi}, \bm{\rho}}))}}{Z_{\bm{\theta}}}]\cr
        &= \mathbb{E}_{\bm{\rho} \sim U(0, 1)}[\mathrm{log} \; q_{\bm{\phi}}(\mathbf{z(\bm{\phi}, \bm{\rho}})|\mathbf{x})] - \mathbb{E}_{\bm{\rho} \sim U(0, 1)}[\mathrm{log} \; \mathrm{e}^{-E_{\bm{\theta}}(\mathbf{z(\bm{\phi}, \bm{\rho}}))}]  + \mathrm{log} \: Z_{\bm{\theta}}\cr
        &= \mathbb{E}_{\bm{\rho} \sim U(0, 1)}[\mathrm{log} \; q_{\bm{\phi}}(\mathbf{z(\bm{\phi}, \bm{\rho}})|\mathbf{x})] + \mathbb{E}_{\bm{\rho} \sim U(0, 1)}[E_{\bm{\theta}}(\mathbf{z(\bm{\phi}, \bm{\rho}}))]  + \mathrm{log} \: Z_{\bm{\theta}}.
    \end{split}
\end{equation}

Since $\nabla_{\bm{\theta}} \, \mathrm{log} \: Z_{\bm{\theta}} = -\nabla_{\bm{\theta}} \, \mathbb{E}_{\mathbf{\tilde{z}}^d \sim p_{\mathbf{\theta}}(\mathbf{\tilde{z}}^d)} [E_{\bm{\theta}}(\mathbf{\tilde{z}}^d)]$ \citep{khoshaman2018gumbolt,song2021train}, unbiased gradients of the KL term with regard to the generative and variational parameters can be obtained as:

\begin{equation} \label{eq:22}
    \nabla_{\bm{\theta}, \bm{\phi}} \{\mathbb{E}_{\bm{\rho} \sim U(0, 1)}[\mathrm{log} \; q_{\bm{\phi}}(\mathbf{z(\bm{\phi}, \bm{\rho}})|\mathbf{x})] + \mathbb{E}_{\bm{\rho} \sim U(0, 1)}[E_{\bm{\theta}}(\mathbf{z(\bm{\phi}, \bm{\rho}}))]  - \mathbb{E}_{\mathbf{\tilde{z}}^d \sim p_{\mathbf{\theta}}(\mathbf{\tilde{z}}^d)}[E_{\bm{\theta}}(\mathbf{\tilde{z}}^d)]\},
\end{equation}

\noindent
where the gradients of the log prior probability are given, as usual, as the difference between a positive and negative phase. The symbolic expression $\mathbf{\tilde{z}}^d$ denotes `fantasy states,' i.e., values of the latent variables produced by the RBM model (prior) distribution, which remain discrete and are not relaxed during training \citep{khoshaman2018gumbolt, vinci2020path}. In the expression above, we have highlighted the fact that the positive-phase energy [$E_{\bm{\theta}}(\mathbf{z(\bm{\phi}, \bm{\rho}}))$] and the negative-phase energy [$E_{\bm{\theta}}(\mathbf{\tilde{z}}^d)$] are calculated [according to the bottom equation of (\ref{eq:6})] using relaxed posterior samples (\textbf{z}) and discrete model samples (fantasy states $\mathbf{\tilde{z}}^d$), respectively. The objective of training is to make the model distribution, $p_{\bm{\theta}}(\mathbf{\tilde{z}}^d)$, as similar as possible to the posterior distribution, $q_{\bm{\phi}}(\mathbf{z}|\mathbf{x})$. We used the persistent-contrastive-divergence (PCD) algorithm \citep{tieleman2008training} to evolve the conditional distributions of the `visible' and `hidden' layers of fantasy states of the negative phase by means of Gibbs sampling, starting from initialization to zero. In PCD, the chains of the fantasy states' values are persistent and continue to evolve over cycles of training (minibatches), without re-initialization at the beginning of the cycle. The PCD algorithm is characterized by short mixing times (fast convergence to the stationary distribution) because the weight updates repel the persistent chains from their current states by raising the energies of the states \citep{hinton2012practical}. It should be noted that this form of training does not require knowledge of the (intractable) partition function $Z_{\bm{\theta}}$. Hence, our VAE model with RBM prior is an energy-based model whose training is based on (unnormalized) prior energies rather than (normalized) prior probabilities.

\subsection{Latent Boltzmann networks} \label{boltzmann}

In a VAE with an RBM prior, the RBM network is located in the latent space; there are no visible RBM units corresponding to input data as in a standalone RBM. Also, by necessity, the latent variables $\mathbf{z}$ of the positive phase are continuous (because they are sampled from the approximate posterior distribution, which is relaxed via the Gumbel-softmax procedure described above during training in order to make the ELBO differentiable). On the other hand, the RBM model samples (fantasy states) $\mathbf{\tilde{z}}^d$ remain discrete variables, as indicated by the superscript, and are not relaxed during training.

\citet{rolfe2016discrete} first developed a DVAE with RBM prior. The model applied the spike-and-exponential transformation to the posterior latents to make them differentiable. \citet{khoshaman2018gumbolt} then modified the DVAE/RBM model by using the Gumbel-softmax trick to bring about the continuous relaxation of the DVAE's latent variables; the authors termed their DVAE model with RBM prior and Gumbel-softmax relaxation `GumBolt.' \citet{vinci2020path} introduced a quantum version of the GumBolt model, based on \citet{amin2018quantum} and \citet{khoshaman2019quantum}. These authors split up the posterior latent units $\mathbf{z}$ into two portions of equal size (denoted as $\mathbf{z}_l$ and $\mathbf{z}_r$ in figure \ref{fig:2}) to implement between them the positive phase of the RBM model according to (\ref{eq:22}) and the energy function given in (\ref{eq:6}). A corresponding approach is taken for the fantasy states $\mathbf{\tilde{z}}^d$ of the negative phase. We designate an RBM model with such variables a `bipartite latent-space RBM.' In such an RBM model, there is no difference in kind between the `visible' and `hidden' units (for example, $\mathbf{z}_l$ and $\mathbf{z}_r$, respectively, in the positive phase). The fantasy states are evolved via PCD Gibbs sampling whereas the posterior latent states remain `clamped' to their original values (VAE encoder output) and are not subjected to Gibbs sampling.

\begin{figure}[h]
    \captionsetup{singlelinecheck = true, justification=justified, font=footnotesize, labelsep=period, width=1\textwidth}
    \centering
    \includegraphics[width=1.0\textwidth]{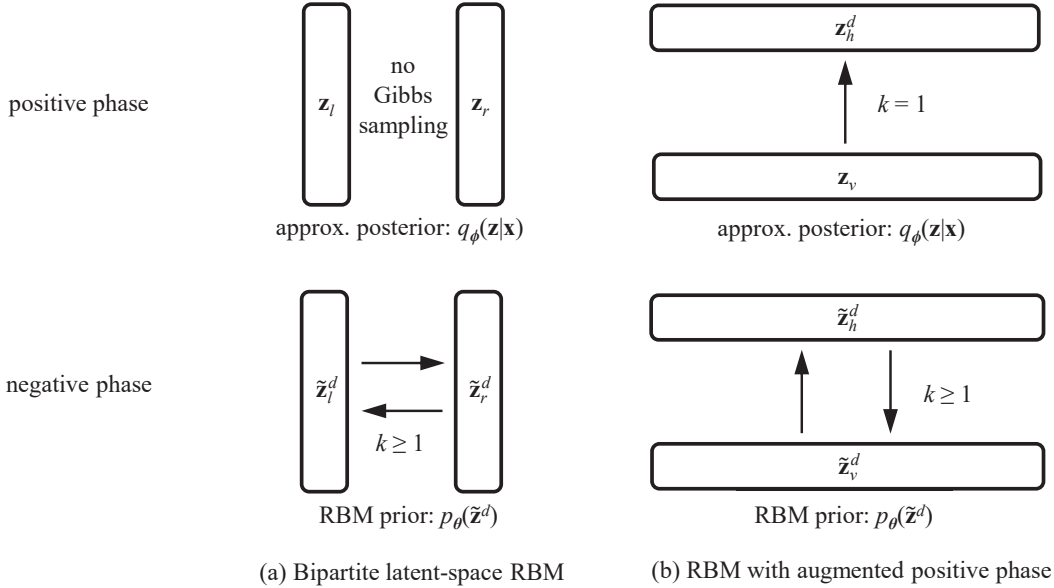}
    \caption{Contrasting architectures of (a) a prior model in which an RBM is implemented between two subdivisions of the VAE's latent space obtained by dividing the latent units $\mathbf{z}$ and $\mathbf{\tilde{z}}^d$, sampled from, respectively, the approximate posterior $q_{\bm{\phi}}(\mathbf{z}|\mathbf{x})$ and model prior $p_{\bm{\theta}}(\mathbf{\tilde{z}}^d)$, into two equally sized parts each (here termed `bipartite latent-space RBM') \citep{rolfe2016discrete, khoshaman2018gumbolt, khoshaman2019quantum, vinci2020path} and of (b) a prior RBM model in which a true hidden layer is added to the layer of posterior latents, which constitute the `visible' layer of the RBM's positive phase  (here termed `RBM with augmented positive phase'). Here, the symbolic expressions for the approximate posterior and prior distributions [$q_{\bm{\phi}}(\mathbf{z}|\mathbf{x})$ and $p_{\bm{\theta}}(\mathbf{\tilde{z}}^d)$, respectively] highlight the fact that the posterior latents are relaxed into continuous random variables \textbf{z} whereas the prior latents (fantasy states) remain discrete variables $\mathbf{\tilde{z}}^d$. In a bipartite latent-space RBM, the latent states \textbf{z} of the approximate posterior remain `clamped' to their original values (VAE encoder output) whereas the fantasy states $\mathbf{\tilde{z}}^d$ of the negative phase are evolved via PCD Gibbs sampling. In this work, we treat the latent variables $\mathbf{z}$ of the approximate posterior as the visible units of the RBM ($\mathbf{z}_v$) and add an equal number of hidden units ($\mathbf{z}_h^d$), thus augmenting the RBM's positive phase. The values of $\mathbf{z}_h^d$ are determined by one-step Gibbs sampling, given the values of $\mathbf{z}_v$. The values of the `visible' and `hidden' units of the negative phase are evolved by PCD Gibbs sampling, as in the bipartite latent-space model. The symbol \textit{k} denotes the number of Gibbs-sampling steps.}
    \addtocounter{figure}{-1}
    \phantomcaption
    \label{fig:2}
\end{figure}

In the studies conducted for this paper, we have adopted a slightly different approach. We consider the posterior latent variables $\mathbf{z}$ to be inputs of the latent-space RBM ($\mathbf{z}_v$) and add an equal number of hidden units ($\mathbf{z}_h^d$) to the model. The values of the hidden units of the positive phase are determined by one-step Gibbs sampling, given the values of the visible units [see top equation of (\ref{eq:7}) for the hidden units' distribution].\footnote[2]{During training, the visible units of the positive phase $\mathbf{z}$, which correspond to the VAE's latent variables, are continuous because they are Bernoulli variables relaxed via the Gumbel-softmax trick to make the objective function differentiable. The fantasy states of the negative phase $\mathbf{\tilde{z}}^d$, by contrast, remain discrete variables and are not relaxed during training. However, the variable type (continuous or discrete) of the hidden units of the positive phase is not obvious. The values of the positive phase's visible and hidden units are used to compute the positive phase's energy $E_{\bm{\theta}}(\mathbf{z}_v, \mathbf{z}_h^{(d)})$, which does not depend on the variational parameters $\bm{\phi}$ [see bottom equation of (\ref{eq:6})]. We evaluated all combinations of continuous and discrete units in the formula for the energy of the positive phase (continuous visible and continuous hidden, continuous visible and discrete hidden, and discrete visible and discrete hidden). The combination of continuous visible units and discrete hidden units produced the best performance, and we chose this combination, based on this empirical observation, as indicated in figure \ref{fig:2}. We will address this question more rigorously in future research.} The values of the `visible' and `hidden' units of the negative phase ($\mathbf{\tilde{z}}_v^d$ and $\mathbf{\tilde{z}}_h^d$, respectively) are evolved by PCD Gibbs sampling, as in the bipartite latent-space model. We call a model with such features an `RBM with augmented positive phase.' The contrasting architectures of the two models are shown in figure \ref{fig:2}. We chose the model shown in figure \ref{fig:2}(b) over the model in (a) to conduct our studies because it had displayed a more dynamic training and slightly better performance in preliminary experiments on the baseline dataset, using otherwise identical hyperparameters, including the number of VAE latents. When the model with augmented positive phase was used as the RBM prior, the mean latent values and mean energies of the positive and negative phases fluctuated more over minibatches of training before convergence to final values than when the bipartite latent-space RBM was used. Similarly, the components of the RBM bias vector \textbf{b} and the elements of the weight matrix \textbf{W} sometimes changed from decreasing to increasing or vice versa while converging to their final values over the epochs of training whereas the parameters of the bipartite latent-space RBM converged to their final values without much intermediary change of direction. The more lively training of the model depicted in figure \ref{fig:2}(b) indicated to us that it might more easily switch between different modes of the system's energy landscape than the faster or more directly converging model illustrated in (a). Lastly, the model in (b) produced slightly better performance metrics (precision, recall, and F1 score) in preliminary test runs on the baseline dataset. We would like to emphasize that we did not comprehensively evaluate and compare the training behavior and performance of the two models because such an assessment and comparison was not the focus of our studies. Both models performed similarly overall, as was to be expected considering their similar structures.

\section{Evaluation of anomaly-detection performance} \label{evaluation}

For the experiments reported in this paper, we trained 16 models (10 for the experiments assessing performance on the baseline dataset with nonoptimal hyperparameters; section \ref{performance}) independently, and we report mean $+/-$ standard deviation. The reconstruction error of the ELBO [negative of left term in (\ref{eq:13})] can be operationalized by the mean squared error (MSE) between the training data and the decoder output (reconstructed training data). We used this error metric when the input data were normalized by mean centering and scaling to unit variance (\textit{z} score), which was the case for the datasets with drop-in-airspeed anomaly during takeoff. The MSE between the reconstructed and original training data is:

\begin{equation} \label{eq:23}
    \mathrm{MSE}_\mathbf{x} = \frac{1}{S} \sum_{s=1}^{S} ||\mathbf{x}^{(s)} - \hat{\mathbf{x}}^{(s)}||_2^2,
\end{equation}

\noindent
where \textbf{x} symbolizes the input data and $\hat{\mathbf{x}}$ their reconstructions and the sum is taken over a minibatch of training data.

We used a different error metric to estimate the reconstruction error when the training data were normalized to lie between zero and one using the transformation

\begin{equation} \label{eq:24}
    \mathbf{x}' = \frac{\mathbf{x} - \mathrm{min}\{\mathbf{x}\}}{\mathrm{max}\{\mathbf{x}\} - \mathrm{min}\{\mathbf{x}\}},
\end{equation}

\noindent
where $\mathbf{x}'$ symbolizes the transformed input data. This was the case for the dataset with a delay-in-flap-deployment anomaly during approach to landing. In this case, we used the binary cross entropy (BCE) to estimate the reconstruction error:

\begin{equation} \label{eq:25}
    \mathrm{BCE}_\mathbf{x} = -\frac{1}{S}(\sum_{s=1}^{S} \sum_{j=1}^{J} x_{j}^{(s)} \; \mathrm{log} \; \hat{x}_{j}^{(s)} + (1 - x_{j}^{(s)}) \; \mathrm{log}(1 - \hat{x}_{j}^{(s)})),
\end{equation}

\noindent
where the sums are over input instances and features. Our experiments showed that the BCE error metric captured the reconstruction error more accurately when the input data were scaled to the interval [0, 1].

Since nominal data points are much more prevalent than anomalous ones, the model primarily learns patterns exhibited by nominal data, and, therefore, their reconstruction errors (MSE or BCE) tend to be smaller than the errors of anomalous points. However, a powerful encoder-decoder model will also fit anomalous data points, which is undesirable when the reconstruction error is used as the metric to identify anomalies. To discourage the fitting to anomalous instances, our models are \textit{variational} (rather than pure) autoencoders. The KL-divergence term in the ELBO of a VAE [(\ref{eq:4})] penalizes the divergence between the approximate posterior $q_{\bm{\phi}}(\mathbf{z}|\mathbf{x})$ and prior $p_{\bm{\theta}}(\mathbf{z})$. We introduce the hyperparameter $\beta$ into the objective function [see (\ref{eq:16})], based on the method developed in \citet{higgins2017beta}, to regulate the relative weighting of the autoencoding and KL terms of the ELBO. When $\beta$ is chosen properly, in such a way that the reduction in KL divergence due to the similarity between posterior and prior outweighs the rise in reconstruction error due to the lack of fit to (the few) anomalous data points, the model can be induced to preferentially fit data in the nominal majority class. Hence, optimal anomaly detection depends on careful tuning of the hyperparameter $\beta$.

Once a VAE model is trained with empirically optimized hyperparameters, we use the reconstruction errors per training instance, \textit{e}, to determine an anomaly-score threshold. The threshold is based on the assumption that the data (nominal and anomalous instances) are normally distributed and is specified as:

\begin{equation} \label{eq:26}
    thr = \langle e \rangle + z \Delta e,
\end{equation}

\noindent
where the angle brackets and $\Delta$ denote the mean and standard deviation, respectively, and \textit{z} the \textit{z} score, derived from the (known) percentage of anomalies in the training set using the quantile function. Once the threshold is determined based on the training data, we identify anomalies in the test data by calculating the anomaly score (reconstruction error) for each test data instance and comparing it to the above threshold; instances with an anomaly score below the threshold are classified as nominal, and instances with a score above the threshold are considered anomalous. Anomaly scores given by the BCE error metric are reasonably normally distributed. However, anomaly scores corresponding to the MSE metric are considerably skewed to the right and possess a long right tail. We applied various normalizing transformations to such anomaly scores, including the square-root, natural-logarithm, and inverse (reciprocal) transformation. On average, the logarithm produced the best anomaly-detection performance, and for this reason, we apply a log transformation to the anomaly scores from these datasets. We sampled both the threshold (based on the training set) and the (log-transformed) anomaly scores (based on the test set) ten times each and used the respective average values to classify data as nominal or anomalous. The thus predicted data labels, determined in an unsupervised way, are then compared with the known true data labels to compute performance metrics.

In all studies, we assume that the nominal data are the negative class and that the anomalous data are the positive class. We assess model performance with three metrics---precision, recall, and F1 score---specified as:

\begin{equation} \label{eq:27}
\begin{split}
        \mathrm{precision} &= \frac{TP}{TP + FP},\cr
        \mathrm{recall} &= \frac{TP}{TP + FN},\cr
        \mathrm{F1 \; score} &= \frac{2}{\mathrm{recall}^{-1} + \mathrm{precision}^{-1}},
    \end{split}
\end{equation}

\noindent
where \textit{TP} represents true positives: correctly identified anomalies, \textit{FP} false positives (alarms): nominal instances incorrectly categorized as anomalous, and \textit{FN} false negatives: missed anomalies or instances that were incorrectly classified as nominal.

\section{Experimental results} \label{results}

All VAE models were implemented in Python using the PyTorch deep-learning library \citep{paszke2019pytorch}. Our VAE model with Gaussian prior is an upgraded version of the CVAE model introduced in \citet{memarzadeh2020unsupervised}, which achieved state-of-the-art performance in detecting anomalies in aviation time series. In all models, the Adam optimizer was used, with a learning rate of $3 \times 10^{-4}$ and default momentum parameters \citep{kingma2014adam}. We used minibatch-based optimization. Minibatches of mutually exclusive training-set instances were re-shuffled for each epoch of training. All minibatches comprised 128 training-set instances, except for the minibatches used for post-training in the transferability study (section \ref{transferability}), which contained 32 instances. We used validation sets to determine combinations of hyperparameter values with good performance. Except for the top layer of the encoder, which outputs the estimated parameters of the approximate posterior and the reparameterized latent variables (and effects the relaxation of discrete latent variables in discrete-variable models), all models use the same encoder and decoder architecture presented in Appendix \hyperref[app:d]{D}.

\subsection{Baseline study: Drop in airspeed during takeoff} \label{baseline}

We determined the baseline performance of the VAE models with Gaussian, Bernoulli, and RBM priors on a dataset of departing flights with drop-in-airspeed anomaly. Subject matter experts ascertained that if the speed of an aircraft drops by more than 20 knots in the first 60 s after becoming airborne, an adverse event might ensue, and, therefore, data points with such a property are classified as anomalous. The dataset contains flight-operations data of commercial flights. It comprises primarily 1-Hz recordings for each flight that cover a variety of systems, including the state, orientation, and speed of the aircraft. The data are acquired in real time aboard the aircraft and downloaded by the airline once the aircraft has reached the destination gate. Each data instance is a 60 s-long recording of seven sensor outputs (five continuous, two discrete) during the initial ascent after becoming airborne. The drop-in-airspeed anomaly is not necessarily the only type of anomaly in the dataset, and the true number of operationally significant anomalies is unknown.

\subsubsection{Model training} \label{training}

For this baseline study, we divided the data into training (60\%), validation (20\%), and test (20\%) sets. We used the training set to train the models, the validation set to monitor overfitting and select hyperparameters, and the test set to assess model performance in an unbiased way. Since the input data were normalized by mean centering and scaling to unit variance, the MSE [(\ref{eq:23})] was used to estimate the reconstruction error. Models were trained for 400 epochs. Model weights tended to converge at about 100 epochs. We did not observe any overfitting with increasing training time, as visualized by the change of the model's loss [negative of the $\beta$-ELBO objective (\ref{eq:16})] over time when evaluated on the validation set (figure \hyperref[fig:1s]{1} in the supplementary material).

We assessed many combinations of hyperparameters, separately for each model type (Gaussian, Bernoulli, RBM), and selected the hyperparameter values that produced the best overall performance in precision, recall, and F1 score. Our approach to use anomaly-detection performance on a validation set, assessed by comparing the data labels predicted by a model with the validation set's known true labels, to tune model hyperparameters is comparable to the hyperparameter-optimization strategies employed in \citet{memarzadeh2020unsupervised} and \citet{li2021autood}. The hyperparameters chosen for each model are given in table \ref{tab:1}. The RBM model, which has a more flexible and expressive prior than the models with standard normal or Bernoulli prior, performed optimally at a lower latent-space dimensionality than the Gaussian and Bernoulli models. We also explored the application of a loss penalty to the RBM coupling weights \textbf{W}, the use of a sampling replay buffer, in which 5\% randomly chosen fantasy states are not determined by the persistent chains but randomly set to 0 or 1 with equal probability \citep{du2019implicit}, KL-term annealing (`warm-up') \citep{sonderby2016ladder}, a hierarchical (conditional) posterior \citep{vahdat2020undirected, vinci2020path}, multiple sampling and averaging of the ELBO and its gradients \citep{vinci2020path}, and the continuous Bernoulli to normalize the ELBO's reconstruction term \citep{loaiza2019continuous}. In the end, we did not apply any of these modifications because none of them improved the performance of our models, where applicable. 

\begin{table}[h]
    \begin{adjustwidth}{1.6cm}{}
        \captionsetup{singlelinecheck = false, justification=justified, font=footnotesize, labelsep=period, skip=0pt}
        \begin{itemize}[noitemsep, nolistsep]
            \caption{\label{tab:1}Hyperparameters used for the Gaussian, Bernoulli, and RBM models.}
            \footnotesize
            \item[]
            \begin{tabular}{@{}+c^c^c^c^c^c^c}
                \toprule
                \rowstyle{\bfseries}
                Model prior&No. latents$^{\rm a}$&beta$^{\rm b}$&lambda$^{\rm c}$&No. fant. part.$^{\rm d}$&Len. pers. chains$^{\rm e}$\cr
                \midrule
                \multicolumn{6}{@{}c}{Drop-in-airspeed anomaly during takeoff}\cr
                \midrule
                Gaussian&256&60&N/A&N/A&N/A\cr
                Bernoulli&128&60&0.1&N/A&N/A\cr 
                RBM&64&60&0.1&500&20\cr
                \midrule
                \multicolumn{6}{@{}c}{Delay-in-flap-deployment anomaly during approach to landing}\cr
                \midrule
                RBM&32&30&0.1&500&25\cr
                \bottomrule
            \end{tabular}
            
            \item[] $^{\rm a}$ Number of latent units.
            \item[] $^{\rm b}$ Hyperparameter $\beta$ controlling the balance between the autoencoding and KL terms.
            \item[] $^{\rm c}$ Temperature of the relaxed Bernoulli distribution.
            \item[] $^{\rm d}$ Number of fantasy particles / persistent chains.
            \item[] $^{\rm e}$ Length of persistent chains.
        \end{itemize}
    \end{adjustwidth}
\end{table}
\normalsize

To assess model performance, the training and validation sets were combined, and the models were re-trained on the combined training/validation set. We then evaluated the performance of the models on the test set. The times of training the Gaussian, Bernoulli, and RBM models on the combined training/validation set for 400 epochs on a Skylake GPU-enhanced node of the Pleiades supercomputer\footnote[3]{\url{https://www.nas.nasa.gov/hecc/resources/pleiades.html}} at the NASA Ames Research Center are shown in table \ref{tab:2}. The Gaussian and Bernoulli models as well as the RBM model with one PCD Gibbs-sampling update during the negative phase require about the same average training time [Gaussian: 1 h 0 min 29 s, Bernoulli: 58 min 27 s, RBM ($k=1$): 1 h 4 min 13 s]. On the other hand, the RBM model with 20 Gibbs updates requires, on average, more time to train [RBM ($k=20$): 1 h 34 min 17 s].

\begin{table}[h]
    \begin{adjustwidth}{4.5cm}{}
        \captionsetup{singlelinecheck = false, justification=justified, font=footnotesize, labelsep=period, skip=-13pt}
        \caption{\label{tab:2}Training times (400 epochs) of VAE models with\\Gaussian, Bernoulli, and RBM priors when trained on the\\combined training/validation set.}
    \end{adjustwidth}
    \begin{adjustwidth}{4.3cm}{}
        \footnotesize
        \item[]\begin{tabular}{l@{\hskip 70 pt}r@{}r}
            \toprule
            \hspace{8 pt}Model&\multicolumn{1}{c}{\hspace{4 pt}Training time}\cr
            \midrule
            Gaussian&1 h 00 min 29 s ($\pm$ 56 s)\cr
            Bernoulli&\hspace{17 pt}58 min 27 s ($\pm$ 19 s)\cr 
            RBM ($k=1$)&1 h 04 min 13 s ($\pm$ 30 s)\cr 
            RBM ($k=20$)&1 h 34 min 17 s ($\pm$ 19 s)\cr 
            \bottomrule
        \end{tabular}
        \item[] Times in parentheses indicate standard deviations.\\The symbol $k$ stands for the number of Gibbs-sampling steps.
    \end{adjustwidth}
\end{table}
\normalsize

Figure \ref{fig:3} shows the values of the latent units of the positive phase [$\mathbf{z} \sim q_{\bm{\phi}}(\mathbf{z}|\mathbf{x})$] and of the negative phase [$\mathbf{\tilde{z}}^d \sim p_{\bm{\theta}}(\mathbf{\tilde{z}}^d)$] averaged over latent dimensions and minibatch instances during a typical training run as well as the corresponding mean energies according to the bottom equation of (\ref{eq:6}), where visible and hidden units reside in the VAE's latent space. Values for each minibatch (171 per epoch) are shown. The figure illustrates that the mean negative-phase values (of the latent variables and energy) closely follow their positive-phase counterparts. However, the mean negative-phase values fluctuate less and are more centered; these findings indicate that the (free-running) negative phase re-produces a smoothed and partially averaged version of the structure of the VAE latent units of the (clamped) positive phase.

\begin{figure}[h]
    \captionsetup{singlelinecheck = true, justification=justified, font=footnotesize, labelsep=period, width=1\textwidth}
    \centering
    \includegraphics[width=1\textwidth]{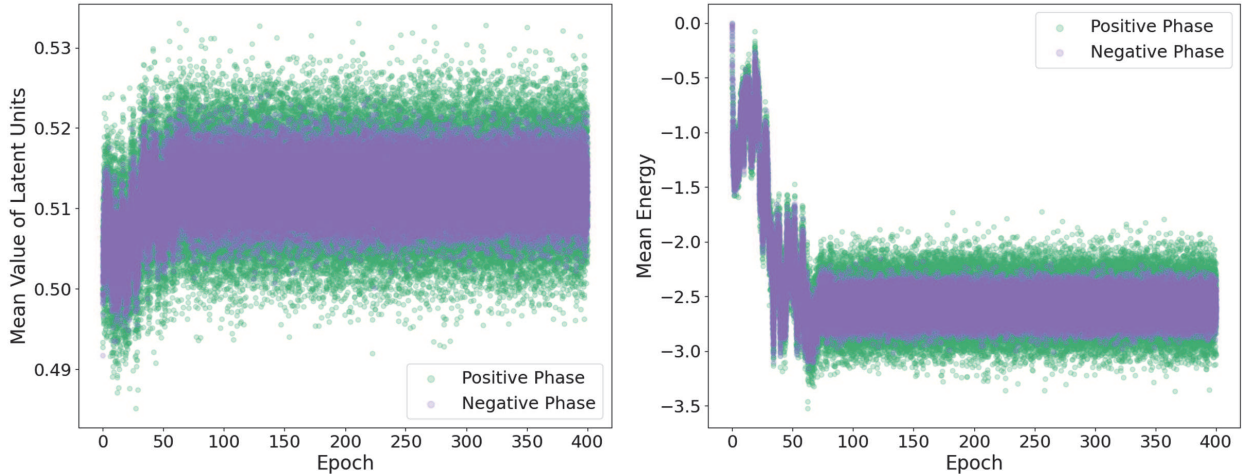}
    \caption{Values of latent units and energy, averaged over latent units and minibatch instances, of the positive phase [$\mathbf{z} \sim q_{\bm{\phi}}(\mathbf{z}|\mathbf{x})$] and of the negative phase [$\mathbf{\tilde{z}}^d \sim p_{\bm{\theta}}(\mathbf{\tilde{z}}^d)$]. The average negative-phase values of these quantities are more centered than the corresponding average values of the positive phase, indicating that the negative phase re-produces a smoothed version of the posterior latents.}
    \addtocounter{figure}{-1}
    \phantomcaption
    \label{fig:3}
\end{figure}

Training of the RBM biases and weights was dynamic, suggesting that the PCD algorithm explored well the energy landscape of the configurations of the system given by the dataset and model (figures \hyperref[fig:2s]{2} and \hyperref[fig:3s]{3} in the supplementary material). Histograms of log-transformed anomaly scores of the RBM model are shown in figure \ref{fig:4} for two training modes. We sampled both the anomaly-score threshold based on the training data as well as the test data's anomaly scores ten times and used the sample statistics to gauge model performance, as described in section \ref{evaluation}. We observed that models enter different modes during training and differ in their anomaly-detection performance depending on the selected mode. The mode with a threshold of about 5.7 for log-transformed anomaly scores produced the best model performance, with F1 scores $>$0.65. Training with modes with a threshold $\gtrsim$5.8, on the other hand, resulted in inferior model performance (F1 score $<$0.65). The superior performance of the mode with $thr \approx 5.7$ is illustrated by the cleaner separation of nominal and anomalous data. Modes with $thr \gtrsim 5.8$ , on the other hand, are characterized by a greater number of false positives (nominal data to the right of the anomaly-score threshold). Other modes, with thresholds between 5.7 and 5.8, were also observed but are less common.

\begin{figure}[h]
    \captionsetup{singlelinecheck = true, justification=justified, font=footnotesize, labelsep=period, width=1\textwidth, belowskip=-5pt}
    \centering
    \includegraphics[width=1\textwidth]{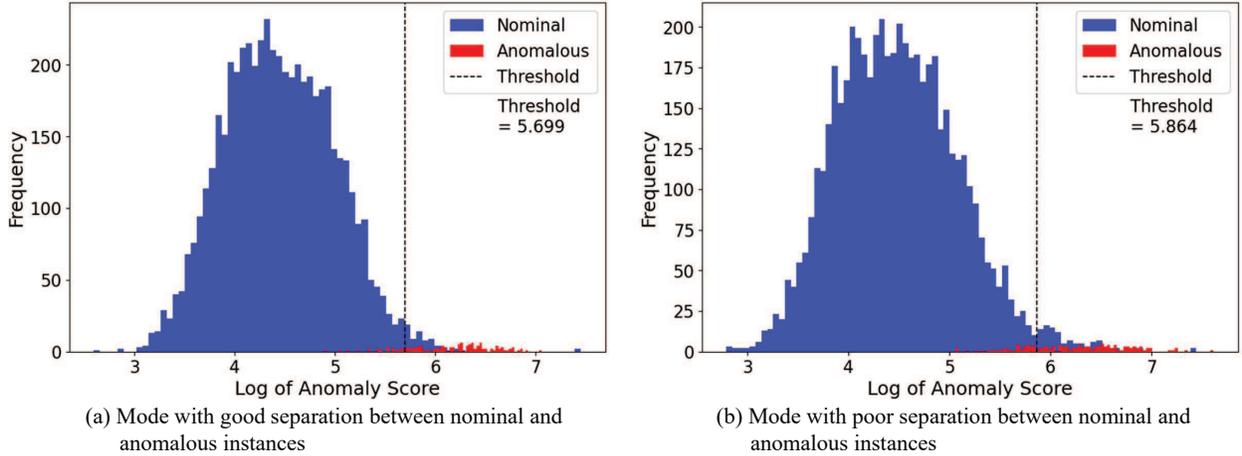}
    \caption{Histograms of average log-transformed anomaly scores for two training modes of the RBM model. Data points to the right of the dashed threshold line are categorized as anomalies. The mode with a threshold of $\sim$5.70 demonstrates a good separation of nominal and anomalous data, whereas the mode with a threshold of $\sim$5.86 exhibits a less good separation, with a relatively high number of nominal data classified as anomalies (false positives).}
    \addtocounter{figure}{-1}
    \phantomcaption
    \label{fig:4}
\end{figure}

\subsubsection{Model performance} \label{performance}

\begin{figure}[!b]
    \captionsetup{singlelinecheck = true, justification=justified, font=footnotesize, labelsep=period, width=1\textwidth}
    \centering
    \includegraphics[width=0.6\textwidth]{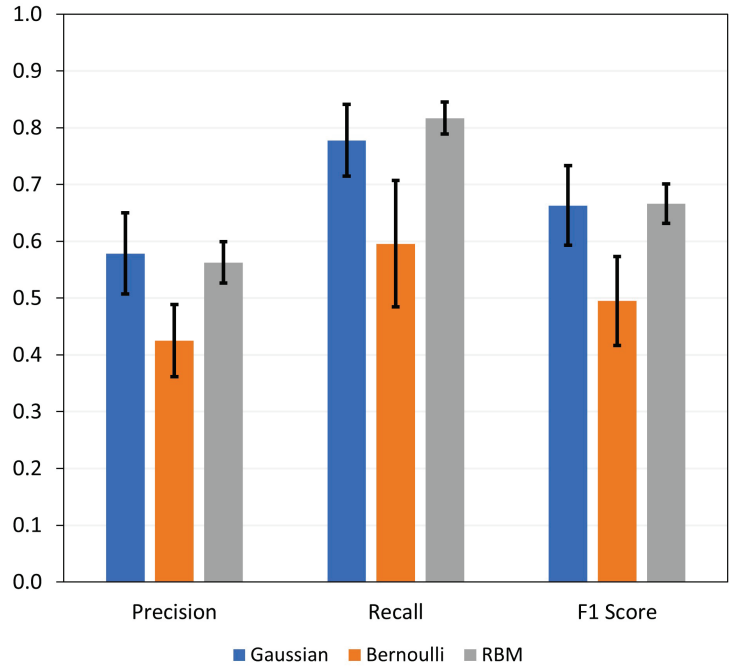}
    \caption{Performance of VAE models with Gaussian, Bernoulli, and RBM priors in the baseline study. Error bars indicate standard deviations of performance metrics among 16 independently trained models. Overall, the models demonstrate excellent performance on this unsupervised anomaly-detection task (average F1 score of 0.608). The more expressive discrete RBM model performs better than the simple discrete Bernoulli model and at a level comparable to the continuous Gaussian model. The DVAE appears to benefit from the flexibility and adaptability of the energy-based RBM prior.}
    \addtocounter{figure}{-1}
    \phantomcaption
    \label{fig:5}
\end{figure}

The performance of the VAE models with Gaussian, Bernoulli, and RBM priors in the baseline study is shown in figure \ref{fig:5}. The RBM model achieved a mean precision of 0.563, a mean recall of 0.817, and a mean F1 score of 0.666. The Gaussian model achieved a similar performance (pr 0.579, rc 0.778, f1 0.663). A notable observation is that the Bernoulli model lags behind both the RBM and the Gaussian model (pr 0.425, rc 0.596, f1 0.495). Our models demonstrate an excellent performance considering that the training was unsupervised and that the similar unsupervised CVAE model with Gaussian prior developed by \citet{memarzadeh2020unsupervised}, which achieved state-of-the-art anomaly-detection performance on aviation datasets, achieved a precision of 0.31 and recall of 0.53 on a dataset similar to the one used in our baseline study.\footnote[4]{The experiment presented in \citet{memarzadeh2020unsupervised} was performed on an extended version of the baseline dataset of departing flights with drop-in-airspeed anomaly that comprised 20 input features, and, in contrast to the models used for the studies conducted for this paper, highly correlated features were routed through five separate encoders and decoders, and the encoder and decoder outputs were combined in, respectively, the latent space and the reconstructed input space. In the experiments reported in both papers, hyperparameters were tuned on validation sets.} Both the Gaussian and Bernoulli models use the simplest factorized priors---the Gaussian model the continuous standard normal distribution and the Bernoulli model the discrete `standard' Bernoulli distribution, with a parameter of 0.5. The models' similar training times also attest to their comparability (see table \ref{tab:2}). Comparing the two models with simple factorized priors, the continuous (Gaussian) model performs better on this dataset, which is dominated by continuous time-series inputs (5 continuous time series, 2 discrete/binary ones). On the other hand, the performance deficit of the simple Bernoulli model is offset by the more expressive RBM model, both of which are discrete VAE models. Precisely the RBM model's flexibility and capability to adapt to the posterior distribution push it to a level of performance on par with the continuous Gaussian model. This observation highlights the performance boost accorded by energy-based modeling and MCMC sampling of the persistent states of the negative phase.

While training and evaluation with optimized hyperparameters allows a fair comparison between models, the approach leads to an overestimation of model performance when no validation set with labeled instances is available, as is frequently the case in practice, including in aeronautics applications. To give an impression of the performance of our models when nonoptimal hyperparameters are used, we trained models with all combinations between four different values for the latent-space dimensionality and five different settings for the hyperparameter $\beta$. The performance of models with 32, 64, 128, and 256 latent units as well as $\beta$ values of 1, 10, 25, 50, and 100 was evaluated. We trained ten Gaussian, Bernoulli, and RBM models each in this way for 300 epochs, which allowed model weights to converge to their final values, and averages of the resultant F1 scores are displayed as heatmaps in figure \ref{fig:6}. The figure demonstrates the performance degradation that occurs when nonoptimal hyperparameters are used. A value of the hyperparameter $\beta$ of 50 or 100 is associated with moderate performance (F1 score between 0.314 and 0.536 across all three models), while a $\beta$ value of 1, 10, or 25 leads to poor performance (F1 score between 0.202 and 0.328). The influence of the hyperparameter $\beta$ on model performance is nonlinear, with a value of 1 or 25 resulting in better performance than a $\beta$ of 10. Performance differences due to different numbers of latent units are less pronounced. All latent-space dimensions investigated in this experiment (32, 64, 128, and 256) produce good performance and correspond to the latent-space dimensions employed in all studies described in this paper (see table \ref{tab:1}).

\begin{figure}[h]
    \captionsetup{singlelinecheck = true, justification=justified, font=footnotesize, labelsep=period, width=1\textwidth}
    \centering
    \includegraphics[width=1\textwidth]{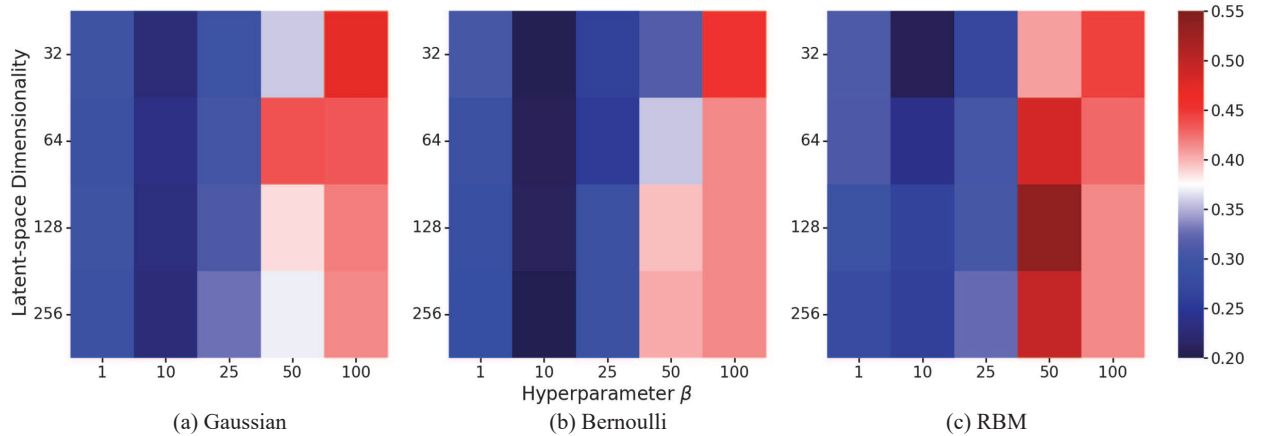}
    \caption{Performance of the Gaussian, Bernoulli, and RBM models in the baseline study after training with nonoptimal hyperparameters. The figure shows heatmaps of average F1 scores for the three models across the hyperparameters latent-space dimensionality and hyperparameter $\beta$, which are common to all three models. Across the three models, F1 scores are noticeably lower than after training with hyperparameters tuned based on validation-set performance, demonstrating the performance degradation incurred by using nonoptimal hyperparameters.}
    \addtocounter{figure}{-1}
    \phantomcaption
    \label{fig:6}
\end{figure}

\subsection{Model transferability} \label{transferability}

The DASHlink dataset\footnote[5]{extracted from the data posted at \url{https://c3.ndc.nasa.gov/dashlink/projects/85/}} used in this experiment consists of time series of sensor data collected during the first 60 s of the takeoff phase of commercial flights. This is a dataset collected independently of the dataset used in the baseline study (section \ref{baseline}) but containing the same input time series as the baseline dataset. The dataset also contains the same anomaly (drop in airspeed during takeoff by more than 20 knots) as the baseline dataset. The transferability study tests the ability of the models tuned and trained on the baseline data to transfer to another dataset containing the same input attributes and anomaly. All transferability experiments were conducted with hyperparameters determined by training on the baseline dataset's training set and assessing model performance on the baseline dataset's validation set (table \ref{tab:1}). We examined two versions of transferability: a strong version of transferability, in which a model trained on the baseline data was directly tested on the DASHlink takeoff data, and a relaxed version, in which a model trained on the baseline data was post-trained on the DASHlink takeoff data for 300 epochs, with model weights initialized to the values acquired after training the model to convergence on the baseline data.

\begin{figure}[h]
    \captionsetup{singlelinecheck = true, justification=justified, font=footnotesize, labelsep=period, width=1\textwidth, belowskip=-10pt}
    \centering
    \includegraphics[width=1\textwidth]{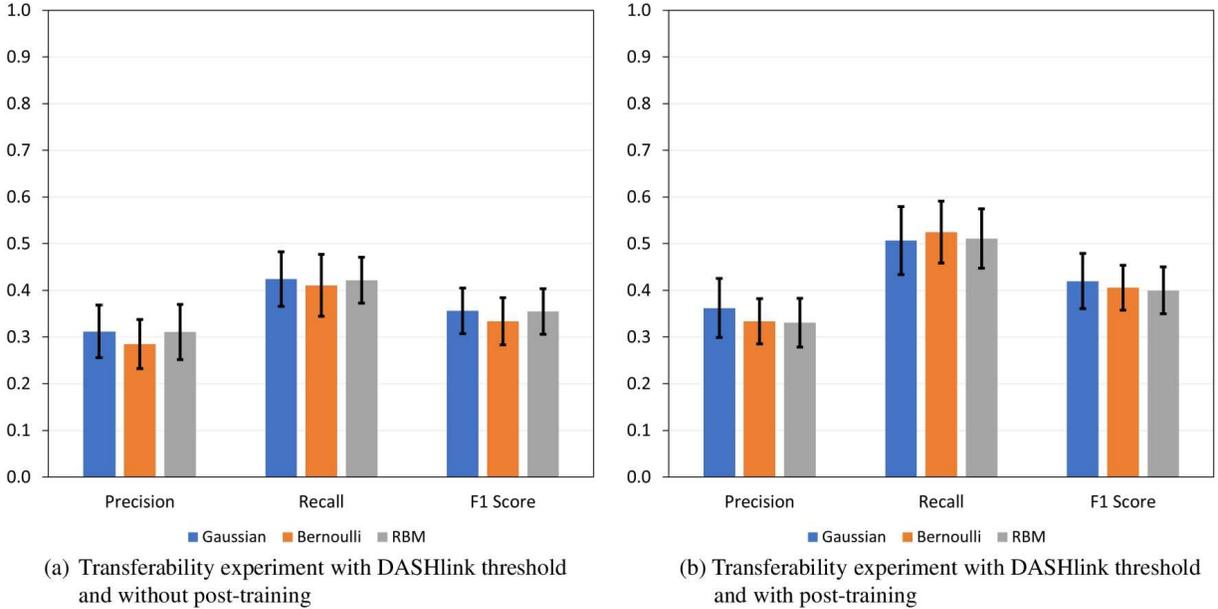}
    \caption{Performance of VAE models with Gaussian, Bernoulli, and RBM priors in the study examining the ability of the models tuned and trained on the baseline data to transfer to the DASHlink takeoff data. Error bars indicate standard deviations of performance metrics among 16 independently trained models. Performance was assessed based on a threshold derived from the DASHlink training set, (a) without or (b) with post-training on the new dataset for 300 epochs with model weights initialized by training to convergence on the baseline dataset. The figure depicts similar performance of all models, trained without supervision, markedly worse than in the baseline experiment, but still in the useful range.}
    \addtocounter{figure}{-1}
    \phantomcaption
    \label{fig:7}
\end{figure}

The DASHlink takeoff dataset was split into training (50\%) and test (50\%) sets. The training set was used to determine the anomaly-score threshold [(\ref{eq:26})] and the test set to compute log anomaly scores [log of (\ref{eq:23})], which were used, in conjunction with the known data labels, to determine model performance. Results are shown in figure \ref{fig:7}. The models with Gaussian, Bernoulli, and RBM priors performed similarly; all results are within each other's error bounds (standard deviations of performance metrics among separately trained models). The average F1 scores of the Gaussian, Bernoulli, and RBM models are 0.356, 0.334, and 0.355, respectively, in the transferability experiment without post-training [figure \ref{fig:7}(a)]. The values of this metric (in the same order) for the experiment with post-training are 0.420, 0.406, and 0.400 [figure \ref{fig:7}(b)]. These results show that post-training on the target dataset slightly improves the overall F1-score average, by 6.03 percentage points. The relatively small extent of the performance improvement highlights the importance of hyperparameters for model performance. Comparing the baseline experiments and the transferability experiments with post-training, the model transfer reduced anomaly-detection performance (as measured by F1 score averaged across models) by 20.0 percentage points. The structure of the results is similar to that in the baseline study, with recall higher than precision. The Bernoulli model does not lag behind the other models, as in the baseline study. Overall, the model transfer markedly reduced model performance while still producing usable results. The transferability results are more difficult to interpret with regard to performance differences between the three models than the baseline results because the experiments from which they were obtained were more complex and involved additional factors (compared with the baseline experiments) since they involved two datasets. We report additional transferability experiments in supplementary section \hyperref[suppl:2]{S2}.

\subsection{Robustness of RBM model: Delay in flap deployment during approach to landing} \label{robustness}

To deepen our insight into the behavior and performance of the VAE model with RBM prior, we conducted a further anomaly-detection study with this model, on another dataset, including \textit{de novo} determination of hyperparameters and training. The experimental approach for this study is similar to the one adopted for the baseline study (section \ref{baseline}). The anomaly in the dataset used for this study is a delay in the deployment of flaps, as judged by a subject matter expert, during the final approach to landing, lasting approximately 160 s.\footnote[6]{The dataset contains $21\,302$ instances, 954 (4.48\%) of which exhibit the delay-in-flap-deployment anomaly. A larger curated unnormalized dataset containing these data is publicly available at \url{https://c3.ndc.nasa.gov/dashlink/resources/1018/}. The dataset used for the study can be reproduced from the posted dataset, up to stochastic differences due to random selection of data points, by (randomly) choosing a subset of the nominal data, choosing only the anomalous data with the pertinent anomaly, scaling the data according to (\ref{eq:24}), with \textbf{x} referring to the training data, and choosing only the attributes Corrected AOA, Barometric Altitude, Computed Airspeed, TE Flap Position, Glideslope Deviation, Core Speed AVG, Pitch Angle, Roll Angle, True Heading, and Wind Speed.} As in the previous experiments, this labeled anomaly is not necessarily the only anomaly in the dataset, and the label is not among the model inputs during the unsupervised training. The dataset contains ten time series of aeronautical sensor outputs, relating to position, orientation, and speed of the aircraft, as well as to the positions of the control surfaces.

The dataset was divided into training (60\%), validation (20\%), and test (20\%) sets. Since the data were normalized to lie in the range of zero to one using (\ref{eq:24}), we used the BCE [(\ref{eq:25})] to estimate the reconstruction error [negative of left term in (\ref{eq:13})]. Training proceeded for 400 epochs and weights converged to their final values after about 100 epochs. We again independently trained 16 models and report the mean +/- standard deviation of their performance. We did not observe any overfitting over the epochs of training, as visualized by the progression of the losses of the validation set. The hyperparameters tried out that produced the best overall performance in precision, recall, and F1 score, evaluated on the validation set, are given in table \ref{tab:1}. To test model performance, the training and validation sets were combined for the purpose of model training, and then the performance of the models trained in this way was evaluated on the test set.  It should be noted that in this study anomaly scores were not log-transformed because the original scores were already approximately normally distributed. Log transformation neither increased the scores' normality nor improved anomaly-detection performance. Additional information is provided in supplementary section \hyperref[suppl:3]{S3}.

\begin{table}[h]
    \begin{adjustwidth}{3.45cm}{}
    \captionsetup{singlelinecheck = false, justification=justified, font=footnotesize, labelsep=period, skip=0pt}
        \caption{\label{tab:3}Performance of the VAE model with RBM prior in the experiment\\with late-deployment-of-flaps anomaly during final approach to landing.}
        \footnotesize
        \item[]\begin{tabular}{c@{\hskip 46 pt}c@{\hskip 46 pt}c}
            \toprule                              
            Performance metric&Average&Standard deviation\cr
            \midrule
            Precision&0.591&0.0543\cr
            Recall&0.647&0.0564\cr 
            F1 Score&0.618&0.0554\cr  
            \bottomrule
        \end{tabular}
    \end{adjustwidth}
\end{table}
\normalsize

Table \ref{tab:3} shows the performance of the RBM model in this case study. The model achieved a mean precision of 0.591, a mean recall of 0.647, and a mean F1 score of 0.618. The mean F1 score is similar to (slightly lower than) the F1 value of 0.666 achieved by the RBM model in the baseline study, which was performed on a dataset with drop-in-airspeed anomaly during takeoff. Consequently, the late-deployment-of-flaps study corroborates the excellent anomaly-detection performance of the unsupervised VAE model with RBM prior observed in the baseline study and illustrates the model's robustness to change of anomaly type and phase of flight.

\section{Discussion} \label{discussion}

\subsection{Model design} \label{design}

We developed two DVAE models. The Bernoulli model employs a factorized Bernoulli distribution as prior, in analogy to the common continuous VAE with Gaussian prior [(\ref{eq:5})]. The RBM model is an attempt to make the DVAE more flexible and expressive. It places an RBM in the VAE's latent space, so that the VAE's latent units are the RBM's inputs, and uses the energy-based restricted Boltzmann distribution given in (\ref{eq:6}) as prior. The RBM model parameters \textbf{a}, \textbf{b}, and \textbf{W} are tuned during training by the interplay between positive- and negative-phase energies, and the `visible' and `hidden' units of the negative phase (fantasy states $\mathbf{\tilde{z}}_v^d$ and $\mathbf{\tilde{z}}_h^d$, respectively) are updated via MCMC Gibbs sampling from (\ref{eq:7}).

DVAE models are less common than VAEs with continuous latent variables because an ELBO objective [(\ref{eq:4})] containing discrete variables cannot be differentiated, thus precluding the computation of ELBO gradients, a necessary operation during the backward pass of training. In order to make DVAEs differentiable, the reparameterization trick \citep{kingma2013auto}, which moves the variational parameters $\bm{\phi}$ from the distribution $q_\phi(\mathbf{z}|\mathbf{x})$ to a more easily differentiable deterministic function $g_{\bm{\phi}}(\bm{\epsilon}, \mathbf{x})$, is extended by a smoothing function \citep{rolfe2016discrete} or a continuous relaxation \citep{maddison2016concrete, jang2016categorical}. Our Bernoulli and RBM models employ the Gumbel-softmax trick \citep{jang2016categorical, maddison2016concrete} to relax the discrete posterior latents $\mathbf{z}^d \sim q_\phi(\mathbf{z}^d|\mathbf{x})$.

Several authors \citep{rolfe2016discrete, khoshaman2018gumbolt, khoshaman2019quantum, vinci2020path} break up the VAE's latent space into two equally sized partitions and implement the positive phase of the RBM between the two halves of latent units $\mathbf{z}$ sampled from the approximate posterior $q_{\bm{\phi}}(\mathbf{z}|\mathbf{x})$ and, similarly, the negative phase between the latent units $\mathbf{\tilde{z}}^d$ sampled from the RBM prior $p_{\bm{\theta}}(\tilde{\mathbf{z}}^d)$. In this paper, we have termed an RBM model with such characteristics a `bipartite latent-space RBM.' In the experiments performed for this paper, we have adopted a somewhat different approach and introduced a hidden layer in the RBM's positive phase, which is obtained in one Gibbs-sampling step from the positive phase's visible layer, comprising the VAE's latent units. The evolution of the negative phase's fantasy states is accomplished by Gibbs sampling in both RBM versions (see figure \ref{fig:2}). We chose this augmented model rather than the bipartite latent-space one to conduct our experiments because it had demonstrated a somewhat more dynamic training and a slightly better performance in preliminary experiments that did not comprehensively evaluate the training behavior and performance characteristics of the two RBM model versions (see discussion at the end of section \ref{discrete}).

In addition to adding a truly hidden layer to the RBM's positive phase, we also introduced the hyperparameter $\beta$ into the ELBO objective function [see (\ref{eq:16})]. While \citet{higgins2017beta} devised the $\beta$-ELBO, in which $\beta$ regulates the trade-off between the autoencoding and KL-divergence terms, as a way to promote disentanglement between latent dimensions by enforcing a minimum similarity between variational posterior $q_{\bm{\phi}}(\mathbf{z}|\mathbf{x})$ and prior $p_{\bm{\theta}}(\mathbf{z})$, we tune $\beta$ to optimize anomaly-detection performance, as measured by performance metrics [(\ref{eq:27})].

\subsection{Model performance} \label{performance1}

We assessed the performance of our VAE models with Gaussian, Bernoulli, and RBM priors in the context of anomaly detection in commercial aeronautics. Prior DVAE models used generation performance on MNIST \citep{lecun1998mnist} and other image datasets to gauge model performance \citep{rolfe2016discrete, khoshaman2018gumbolt, khoshaman2019quantum, vinci2020path}. Generation performance is quantified by the test-set log likelihood or a surrogate such as the ELBO or Q-ELBO. In our anomaly-detection studies, we used the performance metrics precision, recall, and F1 score. The accurate and timely discovery of flight-operations anomalies is important because they can be precursors of potentially serious aviation incidents or accidents. To detect operationally significant anomalies in flight data and preempt future accidents, airlines and transportation agencies will have to increasingly rely on advanced machine-learning techniques applied to historical data or online data streams. Multifactorial and nonlinear anomalies defy traditional anomaly-detection methods, such as exceedance detection \citep{federal2004flight, dillman2015flight}, and the volume and proportion of anomalies characterized by heterogeneous and high-order feature interactions is only expected to increase with increasing airspace complexity, characterized by increasing passenger volume, the integration of unmanned aircraft systems, and urban air mobility. Since labeled data (i.e., data classified as either nominal or anomalous) are costly to obtain and frequently not available and many complex anomalies unknown, unsupervised learning approaches, such as the one portrayed in this paper, are often the preferred or only feasible option.

Both the Gaussian and the Bernoulli models employ simple factorized priors and require similar time to train (table \ref{tab:2}). In these respects, the Bernoulli model is the discrete equivalent of the continuous Gaussian model. The anomaly-detection performance of the three models compares favorably with the performance of a recently developed CVAE and of other unsupervised machine-learning methods, evaluated on a related anomaly-detection task, described in \citet{memarzadeh2020unsupervised}. Our three models achieved F1 scores of $0.663\pm0.0700$ (Gaussian), $0.495\pm0.0785$ (Bernoulli), and $0.666\pm0.0345$ (RBM) when tested on the baseline dataset with drop-in-airspeed anomaly during takeoff. Consequently, the Gaussian and RBM models perform at a similar level, whereas the Bernoulli model falls short in performance. In other words, the performance of the discrete model with simple factorized prior, which is inferior to the capability of the corresponding continuous model, is elevated to the performance of the continuous model by the adoption of a more flexible and expressive energy-based RBM prior that is able to more accurately represent the statistical structure of the input data. Hence, we have implemented an unsupervised discrete deep generative model that performs on par with the analogous continuous generative model that employs a Gaussian prior in detecting anomalies in an aeronautical dataset. The performance decrement due to the usage of nonoptimal hyperparameters was also assessed (figure \ref{fig:6}). Our continuous and discrete models performed similarly in the transferability study (section \ref{transferability}). Our RBM model is ready to be integrated with quantum computing: the discrete (fantasy) states of the negative phase can be obtained by  quantum Boltzmann sampling, by, for example, a quantum annealer or quantum-circuit Born machine (QCBM). On the other hand, a Gaussian VAE, the most common choice in classical deep generative learning, defies a straightforward integration with quantum sampling because it does not sample discrete states from a system whose generative process can be implemented by measuring a parameterizable density operator $\rho_{\bm{\theta}}$ or wave function $\psi_{\bm{\theta}}$ in the computational basis.

The proposed VAE model with RBM prior is robust to change of anomaly type and phase of flight, as demonstrated by its performance on a dataset with late-deployment-of-flaps anomaly during approach to landing, on which it achieved an F1 score of $0.618\pm0.0554$ (section \ref{robustness}). Transfer to a novel dataset without renewed tuning of hyperparameters markedly decreases model performance. The RBM model's F1 score on a dataset on which it was neither tuned nor trained but which contained the same input data and anomaly and spanned the same phase of flight was $0.355\pm0.0486$, 31.1 percentage points lower than the performance on the dataset from which the transfer occurred. While such a drop represents a significant deterioration in performance, the post-transfer performance is still respectable, considering that the datasets involved consist of multiple complex time series and the training was executed without supervision. Transferability was mildly improved by post-training on the target dataset: initializing model weights to the (converged) values obtained at the end of training on the original dataset, post-training on the new dataset for 300 epochs raised the RBM model's average F1 score by 4.51 percentage points to a value of 0.400.

Latent-space dimensionality and the hyperparameter $\beta$ are hyperparameters that have a strong effect on model performance (in all three models) and require optimization for each application. Models with 32--256 latent units generally exhibited good performance, whereas model performance was very sensitive to the specific choice of $\beta$; hence, the choice of the value of the hyperparameter $\beta$ is application-specific. The temperature $\lambda$ of the concrete distribution \citep{maddison2016concrete, jang2016categorical}, which determines the bias introduced by the continuous Gumbel-softmax relaxation and the differentiability of the model parameters, is also important for the discrete models (Bernoulli and RBM) and requires tuning. We used a $\lambda$ of 0.1 throughout our studies because it balances the trade-off between estimation bias and differentiability and had proved the optimal value among the ones tested. We did not use an annealing schedule to gradually reduce $\lambda$ over the epochs of training, but we used discrete (`hard') Bernoulli variables for validation and testing. While the length of the persistent chains (fantasy particles) had a mild influence on the performance of the RBM model, the number of fantasy particles was quite unimportant, due to the averaging of the particles' (negative-phase) energies at the end of each series of Gibbs updates (per minibatch); we used 500 throughout. When optimal, or even adequate, hyperparameters are unknown and labeled data not available, which is often the case in practical applications, the performance of unsupervised anomaly-detection models can frequently be improved by making the effort to prepare a small labeled validation set that includes relevant anomalies and using it for hyperparameter tuning \citep{soenen2021effect, antoniadis2022systematic}. 

Mean negative-phase values of the latent variables and energy tightly followed the corresponding values of the positive phase, but were more centered, suggesting that the free-running negative phase re-produces a smoothed version of the clamped positive phase (figure \ref{fig:3}). We were able to boost anomaly-detection performance by applying a normalizing transformation to skewed anomaly scores (MSEs) with conspicuous right tail. We always used a log transformation, but less skewed scores might benefit more from a square-root transformation and for more strongly skewed scores an inverse transformation might be optimal. We would like to note that our VAE models with continuous and discrete priors are universal generative models and not limited in application to aeronautics data, but, with slight modification, can also be applied to other time-series data.

\section{Conclusions} \label{conclusions}

We developed unsupervised VAE models with convolutional encoder and decoder layers and a latent space based on Gaussian, Bernoulli, and RBM priors. The discrete Bernoulli and RBM versions are an attempt to design models whose latent space captures the discrete nature of objects processed by machine-learning models, such as semantic classes, causal factors, digital samples, and other discrete entities. The RBM model allows a straightforward integration with quantum computing because the fantasy states of the RBM's negative phase can be obtained by measuring the states of a quantum generative process, such as a quantum Boltzmann machine (QBM) implemented on an annealer or a QCBM on a circuit. Our objective is the maximization of the $\beta$-ELBO, in which the hyperparameter $\beta$ regulates the trade-off between reconstruction fidelity and regularization via a prior distribution placed on the latent variables.

We tested our VAE models on datasets comprising time series of commercial flights that contain anomalies during the takeoff or approach-to-landing phases of flight. It is expected that the use of unsupervised and semi-supervised machine-learning techniques to identify anomalies in aviation is going to increase because of the high cost of labeling instances manually, the limitations associated with the identification of anomalies based on simple heuristics, and the ever increasing airspace complexity over the course of the 21st century. While the estimated generative log likelihood or a related quantity is used to gauge generative-model performance in re-generation studies (e.g., of imagery), anomaly detection relies on metrics such as precision, recall, and F1 score to assess model performance.

While all unsupervised models exhibit good performance overall, the discrete Bernoulli model performs more poorly than the other two. However, the more expressive RBM model, which, during training, employs unnormalized energies rather than probabilities and MCMC to sample from the prior distribution, is a discrete-variable model whose performance matches that of the continuous Gaussian model. Our models are robust to changes in type of anomaly and phase of flight. Transferring a model without re-tuning of hyperparameters or re-training to a new dataset with the same anomaly results in an anomaly-detection performance that is markedly impaired, but still acceptable, considering the nature of the data and training.

In future studies, we will use more advanced algorithms, such as parallel tempering and adaptive-step PCD, to conduct negative-phase sampling, to see if such schemes improve model performance. We will also devise an estimator of the log partition function, to compute log-likelihood estimates for future generation studies. We also intend to use conditional VAEs or other deep generative models to generate artificial anomalies, to enhance anomaly-detection datasets and, ultimately, the performance of anomaly-detection algorithms. We plan to integrate quantum capabilities by developing a QBM, implemented by quantum annealing or QCBM. In addition, we intend to increase the expressiveness of our prior model by replacing the relatively simple RBM network with a more sophisticated energy-based feedforward network, and we will attempt to integrate such an EBM with quantum computing.

\section*{Acknowledgments} \label{acknowledgments}

We would like to thank Bryan Matthews for his help with preparing the datasets used in the studies reported in this paper and Arash Vahdat for a helpful discussion while preparing this work. This work was supported by NASA Ames Center Innovation Funding.  A.A.A, M.M., and P.A.L. acknowledge funding support from NASA Academic Mission Services contract NNA16BD14C (System-Wide Safety project). A.A. acknowledges funding support from NASA Academic Mission Services contract NNA16BD14C and from NSF award CCF-1918549 through the Feynman Quantum Academy Internship Program.

\begin{appendices}

\renewcommand{\theequation}{A\arabic{equation}}
\renewcommand{\figurename}{Figure A}

\section*{Appendix A. Unbiased gradients with respect to generative and variational parameters of VAE with factorized Gaussian prior} \label{app:a}

Unbiased gradients with respect to the generative parameters $\bm{\theta}$ are straightforward to obtain, and we can write:

\setcounter{equation}{0}
\begin{equation} \label{eq:9}
    \nabla_{\bm{\theta}} \mathbb{E}_{\mathbf{z} \sim q_{\bm{\phi}}(\mathbf{z}|\mathbf{x})}[f(\mathbf{z})] = \mathbb{E}_{\mathbf{z} \sim q_{\bm{\phi}}(\mathbf{z}|\mathbf{x})}[\nabla_{\bm{\theta}} f(\mathbf{z})] \simeq \frac{1}{S} \sum_{s=1}^{S} \nabla_{\bm{\theta}} f(\mathbf{z}^{(s)}).
\end{equation}

\noindent
The mean on the right of (\ref{eq:9}) is a Monte Carlo estimate of the gradient with respect to $\bm{\theta}$, in which $\mathbf{z}^{(s)}$ is an i.i.d. sample from $q_{\bm{\phi}}(\mathbf{z}|\mathbf{x})$ and $S$ indicates the size of the minibatch.

Unbiased gradients with respect to the variational parameters $\bm{\phi}$ are more difficult to obtain because the expectations in (\ref{eq:4}) are estimated by sampling from a probability distribution that depends on $\bm{\phi}$ \citep{khoshaman2019quantum}. Based on the gradient of the logarithm of $q_{\bm{\phi}}$, $\nabla_{\bm{\phi}} \, \mathrm{log} \: q_{\bm{\phi}}$, the score-function estimator \citep{fu2006gradient, williams1992simple, glynn1990likelihood} calculates gradients with respect to $\bm{\phi}$ as:

\begin{equation} \label{eq:10}
    \nabla_{\bm{\phi}} \mathbb{E}_{\mathbf{z} \sim q_{\bm{\phi}}(\mathbf{z}|\mathbf{x})}[f(\mathbf{z})] = \mathbb{E}_{\mathbf{z} \sim q_{\bm{\phi}}(\mathbf{z}|\mathbf{x})}[f(\mathbf{z}) \nabla_{\bm{\phi}} \mathrm{log} \: q_{\bm{\phi}}(\mathbf{z}|\mathbf{x})] \simeq \frac{1}{S} \sum_{s=1}^{S} f(\mathbf{z}^{(s)}) \nabla_{\bm{\phi}} \mathrm{log} \: q_{\bm{\phi}}(\mathbf{z}^{(s)}|\mathbf{x})).
\end{equation}

\noindent
Here, we assumed for simplicity that $f$ does not depend on $\bm{\phi}$. Unfortunately, gradients based on the score-function estimator are characterized by high variance and require the use of intricate variance-reduction techniques (such as control variates) in practical applications \citep{mnih2014neural, grathwohl2017backpropagation}.

The reparameterization trick \citep{kingma2013auto} is used in VAEs as a low-variance alternative to the score-function estimator. Here, the random variable $\mathbf{z} \sim q_{\bm{\phi}}(\mathbf{z}|\mathbf{x})$ is re-expressed by means of an auxiliary random variable $\bm{\epsilon} \sim p(\bm{\epsilon})$ independent of $\bm{\phi}$ and a deterministic function $g_{\bm{\phi}}(\cdot)$ as $\mathbf{z} = g_{\bm{\phi}}(\bm{\epsilon}, \mathbf{x})$. We then can write $\mathbb{E}_{\mathbf{z} \sim q_{\bm{\phi}}(\mathbf{z}|\mathbf{x})}[f(\mathbf{z})] = \mathbb{E}_{\mathbf{\epsilon} \sim p(\bm{\epsilon})}[f(g_{\bm{\phi}}(\bm{\epsilon}, \mathbf{x}))]$ and obtain unbiased Monte Carlo estimates of the gradient with respect to $\bm{\phi}$ by moving the gradient into the expectation:

\begin{equation} \label{eq:11}
    \nabla_{\bm{\phi}} \mathbb{E}_{\mathbf{z} \sim q_{\bm{\phi}}(\mathbf{z}|\mathbf{x})}[f(\mathbf{z})] = \mathbb{E}_{\mathbf{\epsilon} \sim p(\mathbf{\epsilon})}[\nabla_{\bm{\phi}} f(g_{\bm{\phi}}(\bm{\epsilon}, \mathbf{x}))] \simeq \frac{1}{S} \sum_{s=1}^{S} \nabla_{\bm{\phi}} f(g_{\bm{\phi}}(\bm{\epsilon}^{(s)}, \mathbf{x})),
\end{equation}

\noindent
where $\bm{\epsilon}^{(s)} \sim p(\bm{\epsilon})$. The reparameterization trick transfers the dependence on $\bm{\phi}$ from $q_{\bm{\phi}}$ into $f$, substituting the problem of estimating the gradient with respect to the variational parameters of a distribution with the simpler problem of estimating the gradient with respect to the variational parameters of a deterministic function \citep{maddison2016concrete}.

In a VAE with a factorized Gaussian prior, the latent variables produced by the encoder $q_{\bm{\phi}}(\mathbf{z}|\mathbf{x}) = \mathcal{N}(\mathbf{z}; \bm{\mu}, \mathrm{diag}(\bm{\sigma}^2)) \equiv \prod_l \mathcal{N}(z_l; \mu_l, \sigma_l^2))$ can be reparameterized as $\mathbf{z} = \bm{\mu} + \bm{\sigma} \odot \bm{\epsilon}$, where $\odot$ represents the elementwise product and $\bm{\epsilon} \sim \mathcal{N}(\mathbf{0}, \mathbf{I})$ \citep{kingma2019introduction}. In other words, the components of the reparameterized latent vector $\mathbf{z}$ are reparameterized univariate Gaussians $z_l = \mu_l + \sigma_l \epsilon_l$, $\epsilon_l \sim \mathcal{N}(0, 1)$. Unbiased gradients with respect to the variational parameters $\bm{\phi}$ are then obtained as:

\begin{equation} \label{eq:12}
    \nabla_{\bm{\phi}} \mathbb{E}_{\mathbf{z} \sim \mathcal{N}(\bm{\mu}, \mathrm{diag}(\bm{\sigma}^2))}[f(\mathbf{z})] = \mathbb{E}_{\mathbf{\epsilon} \sim \mathcal{N}(\mathbf{0}, \mathbf{I})}[\nabla_{\bm{\phi}} f(\bm{\mu} + \bm{\sigma} \odot \bm{\epsilon})] \simeq \frac{1}{S} \sum_{s=1}^{S} \nabla_{\bm{\phi}} f(\bm{\mu} + \bm{\sigma} \odot \bm{\epsilon}^{(s)}).
\end{equation}

\section*{Appendix B. Derivation of $\bm{\beta}$-ELBO} \label{app:b}

The constrained optimization problem for a VAE is specified in (\ref{eq:14}), where $\delta$ controls the strength of the applied constraint.

\begin{equation} \label{eq:14}
    \max_{\bm{\theta}, \bm{\phi}} \mathbb{E}_{\mathbf{z} \sim q_{\bm{\phi}}(\mathbf{z}|\mathbf{x})}[\mathrm{log} \; p_{\bm{\theta}}(\mathbf{x}|\mathbf{z})]\mathrm{s.t.} \quad D_{\mathrm{KL}}(q_{\bm{\phi}}(\mathbf{z}|\mathbf{x})||p_{\bm{\theta}}(\mathbf{z})) < \delta
\end{equation}

Using KKT conditions \citep{kuhn1951nonlinear, karush1939minima} to re-write (\ref{eq:14}) as a Lagrangian yields:

\begin{equation} \label{eq:15}
    \mathcal{F}(\bm{\theta}, \bm{\phi}, \beta;\mathbf{x}) = \mathbb{E}_{\mathbf{z} \sim q_{\bm{\phi}}(\mathbf{z}|\mathbf{x})}[\mathrm{log} \; p_{\bm{\theta}}(\mathbf{x}|\mathbf{z})] - \beta(D_{\mathrm{KL}}(q_{\bm{\phi}}(\mathbf{z}|\mathbf{x})||p_{\bm{\theta}}(\mathbf{z})) - \delta),
\end{equation}

\noindent
where the KKT multiplier $\beta$ is a regularization coefficient that constrains latent-space capacity and exerts implicit pressure on the latent-space variables $\mathbf{z}$, which represent the input data $\mathbf{x}$, to become less correlated by drawing each component variable $z_l$ in the direction of the corresponding variable sampled from the prior. \citet{higgins2017beta} demonstrate that a higher $\beta$ leads to less entangled latent variables, but it also decreases reconstruction quality. Disentanglement is easy to visualize in images, by observing the continuous change of a factor when a latent dimension is varied while the others are held constant. By eliminating the $\delta$ term, (\ref{eq:15}) can be written as a $\beta$-ELBO:

\begin{equation} \label{eq:16a}
    \mathcal{L}(\bm{\theta}, \bm{\phi};\mathbf{x}, \beta) = \mathbb{E}_{\mathbf{z} \sim q_{\bm{\phi}}(\mathbf{z}|\mathbf{x})}[\mathrm{log} \; p_{\bm{\theta}}(\mathbf{x}|\mathbf{z})] - \beta \, D_{\mathrm{KL}}(q_{\bm{\phi}}(\mathbf{z}|\mathbf{x})||p_{\bm{\theta}}(\mathbf{z})).
\end{equation}

Since both $\beta$ and $\delta$ are non-negative constants in (\ref{eq:15}), $\mathcal{L}$ bounds $\mathcal{F}$ from below: $\mathcal{F}(\bm{\theta}, \bm{\phi}, \beta; \mathbf{x}) \geq \mathcal{L}(\bm{\theta}, \bm{\phi};\mathbf{x}, \beta)$.

\section*{Appendix C. Analytic expression for KL-divergence term in ELBO objective of VAE with Bernoulli prior} \label{app:c}

For a Bernoulli prior, the KL term can also be written as:

\begin{equation} \label{eq:19}
    D_{\mathrm{KL}}(\mathcal{B}(\mathbf{z}^d; \mathbf{q})||\mathcal{B}(\mathbf{z}^d; \mathbf{0.5})) = \sum_{l=1}^{L} \sum_{k=0}^{1} q_l^{k} \; (1-q_l)^{1-k} \; \mathrm{log} \frac{q_l^{k} \; (1-q_l)^{1-k}}{0.5^{k} (1-0.5)^{1-k}},
\end{equation}

\noindent
where the possible outcomes $k$ in the support of the posterior distribution $q_{\bm{\phi}}(\mathbf{z}^d|\mathbf{x})$ are considered to be \textit{discrete} and $l$ indexes a latent variable. Algebraic manipulation of (\ref{eq:19}) then leads to an alternative analytic expression for the KL term of a VAE with Bernoulli prior:

\begin{equation} \label{eq:20}
    D_{\mathrm{KL}}(\mathcal{B}(\mathbf{z}^d; \bm{\alpha}^{\mathbf{q}})||\mathcal{B}(\mathbf{z}^d; \mathbf{1})) = \sum_{l=1}^{L} \frac{\alpha_l^q \; \mathrm{log}(\frac{\alpha_l^q}{\alpha_l^q+1}) + \mathrm{log}(\frac{1}{\alpha_l^q+1}) + (\alpha_l^q+1) \, \mathrm{log} \: 2} {\alpha_l^q+1},
\end{equation}

\noindent
where $\bm{\alpha}^{\mathbf{q}}$ stands for the vector of the odds of the parameters of the Bernoulli posterior and reparameterizes it. We tested this analytic method using the odds of the \textit{relaxed} Bernoulli posterior to compute the KL-divergence term in the $\beta$-ELBO loss [negative of (\ref{eq:16})] during training, and it basically produces the same results as the stochastic approximation described in section \ref{discrete} (same results within the bounds of stochastic variability).

\section*{Appendix D. Architecture of $\bm{\beta}$-CVAE models} \label{app:d}

Figure \hyperref[fig:a1]{A1} shows the architecture of the encoder of the VAE models. The input data traverse three parallel branches of 1D convolution operations with different filter sizes (first numeric) and kernel sizes (second numeric), followed by batch normalization and ReLU activation, and finally max pooling of size 2. The decoder is identical to an inverted encoder in which 1D convolutions have been replaced with 1D transpose convolutions and max pooling with upsampling (with bilinear interpolation).

\setcounter{figure}{0} 
\begin{figure}[h]
    \captionsetup{singlelinecheck = true, justification=justified, font=footnotesize, labelsep=period, width=0.8\textwidth}
    \centering
    \includegraphics[width=0.8\textwidth]{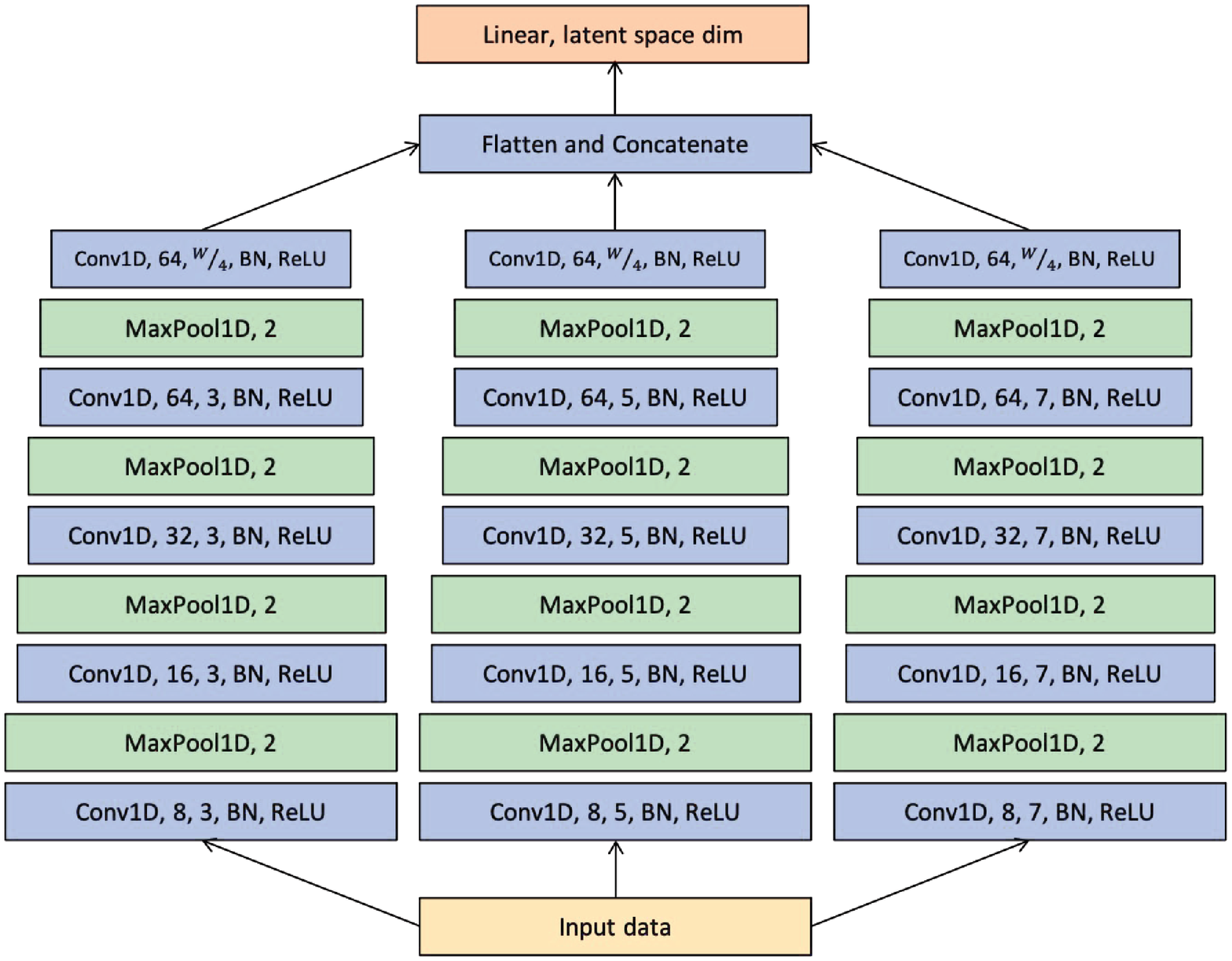}
    \let\nobreakspace\relax
    \caption{Architecture of the encoder of the VAE models.}
    \addtocounter{figure}{-1}
    \phantomcaption
    \label{fig:a1}
\end{figure}

\end{appendices}

\bibliography{References} \label{bibliography}

\end{document}


\interfootnotelinepenalty=10000

\section*{Supplementary material} \label{supplement}

\subsection*{S1. The necessity of discrete-variable models} \label{suppl:1}

Real-world objects are frequently discrete: the objects in an image; phonemes in speech; symbols, words, and sentences in language. Our study on the unsupervised detection of anomalies in aeronautics data requires the differentiation of discrete classes (nominal vs. anomalous instances), and the input to our models consists of multivariate time series consisting of discrete temporal data points. Consequently, internal discrete-variable representations are frequently more interpretable and a more natural fit to modalities of interest present in datasets \citep{chen2016infogan, van2017neural}. When causes and/or consequences are discrete (`If the store doesn't have peaches, I'll buy apples.'), they are also a natural fit for complex reasoning, planning, and predictive learning \citep{van2017neural}.

Discrete variables also naturally represent typical machine-learning operations, such as choosing between models or variables \citep{khoshaman2018gumbolt}. An unsupervised model with a latent space composed of discrete variables can learn to disentangle content and style information of images \citep{makhzani2017pixelgan}. In semi-supervised learning, discrete variables can be used to label distinct semantic classes and to train generative models with more meaningful representations \citep{kingma2014semi, maaloe2017semi, khoshaman2019quantum}. Moreover, discrete variables pose a low burden on computer memory and are computationally efficient \citep{rae2016scaling, jang2016categorical}; they are also a natural match to digital computation.

\citet{rolfe2016discrete} points out that many practically relevant datasets contain information on discrete objects subject to continuous transformation, such as image datasets of natural objects that change position and orientation and experience variations in scene illumination. Such datasets comprise multiple disconnected smooth manifolds. The author notices that while continuous variations in pose and illumination are typical natural phenomena, it is very difficult to transform the image of a person to that of a car while remaining on the manifold of natural images. Rolfe goes on to suggest to model the selection of discrete real-world objects using discrete variables and the continuous transformations applied to each disconnected component with continuous variables. The author implements such a framework using an autoregressive network consisting of layers of continuous latent variables conditioned on a layer of discrete variables. Our models of discrete convolutional VAEs also follow Rolfe's prescription by combining encoders and decoders with continuous convolutional layers with a latent space with discrete variables.

While discrete-variable models promise gains in performance and improvements in efficiency in classical computing, they are outright obligatory for quantum computing. Quantum variables take the form of density matrices, the measurements of which will produce discrete values (bitstrings) as projective measurements of qubits $i = 1, 2, ..., n$ in the computational basis produce values $m_i \in \{-1, +1\}.$ The physical correlates of logical qubits are two-level quantum systems such as the polarization direction of light, Fock states of bosons, quadratures of coherent states, the spin direction of atomic particles, or the flow direction of superconducting current \citep{nielsen2000quantum, nakahara2008quantum, knill2010quantum}. Consequently, the measured binary qubit values $m_i$ constitute the interface between the quantum and classical computing worlds; their processing by the classical component of quantum-classical hybrid models requires discrete variables.

\subsection*{S2. Additional experiments relating to section \hyperref[transferability]{5.2 Model transferability}} \label{suppl:2}

We show the performance of two additional transferability experiments in supplementary figure \ref{fig:4s}. The data differ from the data shown in figure \hyperref[fig:7]{7} in that the threshold for the data shown in supplementary figure \ref{fig:4s}(a) is based on the training set of the baseline dataset and the threshold for the data depicted in supplementary figure \ref{fig:4s}(b) is derived from an average of the training-set thresholds of the baseline and DASHlink takeoff datasets. We used this mixed threshold because in the transferability experiments with post-training both the baseline and the DASHlink takeoff datasets were involved in the training of the models. The results of the experiment with baseline threshold and without post-training are quite imbalanced in the sense that recall is much better than precision, and the overall average F1 score is 3.74 percentage points below the one in the experiment with DASHlink/takeoff threshold and without post-training [figure \hyperref[fig:7]{7(a)}]. The results of the experiment with mixed baseline-DASHlink/takeoff threshold and post-training are very similar to the ones of the post-training experiment with DASHlink threshold.

\subsection*{S3. Additional information relating to section \hyperref[robustness]{5.3 Robustness of RBM model: Delay in flap deployment during approach to landing}} \label{suppl:3}

Histograms of anomaly scores for two training modes of the RBM model, with anomaly-score thresholds of about 980 and 1011, are shown in supplementary figure \ref{fig:5s}. In this study, anomaly scores correspond to the BCE [(\hyperref[eq:25]{17})], which contains a log transformation of the reconstructed data, and the greater normality of the raw anomaly scores is presumably due to the BCE error metric; in the studies described in sections \hyperref[baseline]{5.1} and \hyperref[transferability]{5.2}, we applied the MSE error metric to \textit{z}-transformed reconstructed and original input data and, during evaluation of model performance, log-transformed the resultant right-skewed anomaly scores. As in the baseline study with a dataset containing a drop-in-airspeed anomaly (see section \hyperref[baseline]{5.1}), we observed that models enter different modes during training and differ in their anomaly-detection performance depending on the selected mode. The mode with an anomaly-score threshold of about 980 produced a good model performance, with F1 scores $\sim$0.64, whereas the mode with a threshold of about 1011 resulted in inferior model performance (F1 scores $\sim$0.42). The superior performance of the mode with $thr \approx 980$ is illustrated by the cleaner separation of nominal and anomalous data. The mode with $thr \approx 1011$, on the other hand, is characterized by a greater number of false negatives (anomalous data to the left of the anomaly-score threshold). Other modes were not observed in this experiment.

\renewcommand{\figurename}{Figure S}
\subsection*{S4. Additional figures} \label{suppl:4}

\setcounter{figure}{0} 
\begin{figure}[!htb]
    \captionsetup{singlelinecheck = true, justification=justified, font=footnotesize, labelsep=period, width=1\textwidth}
    \centering
    \includegraphics[width=1\textwidth]{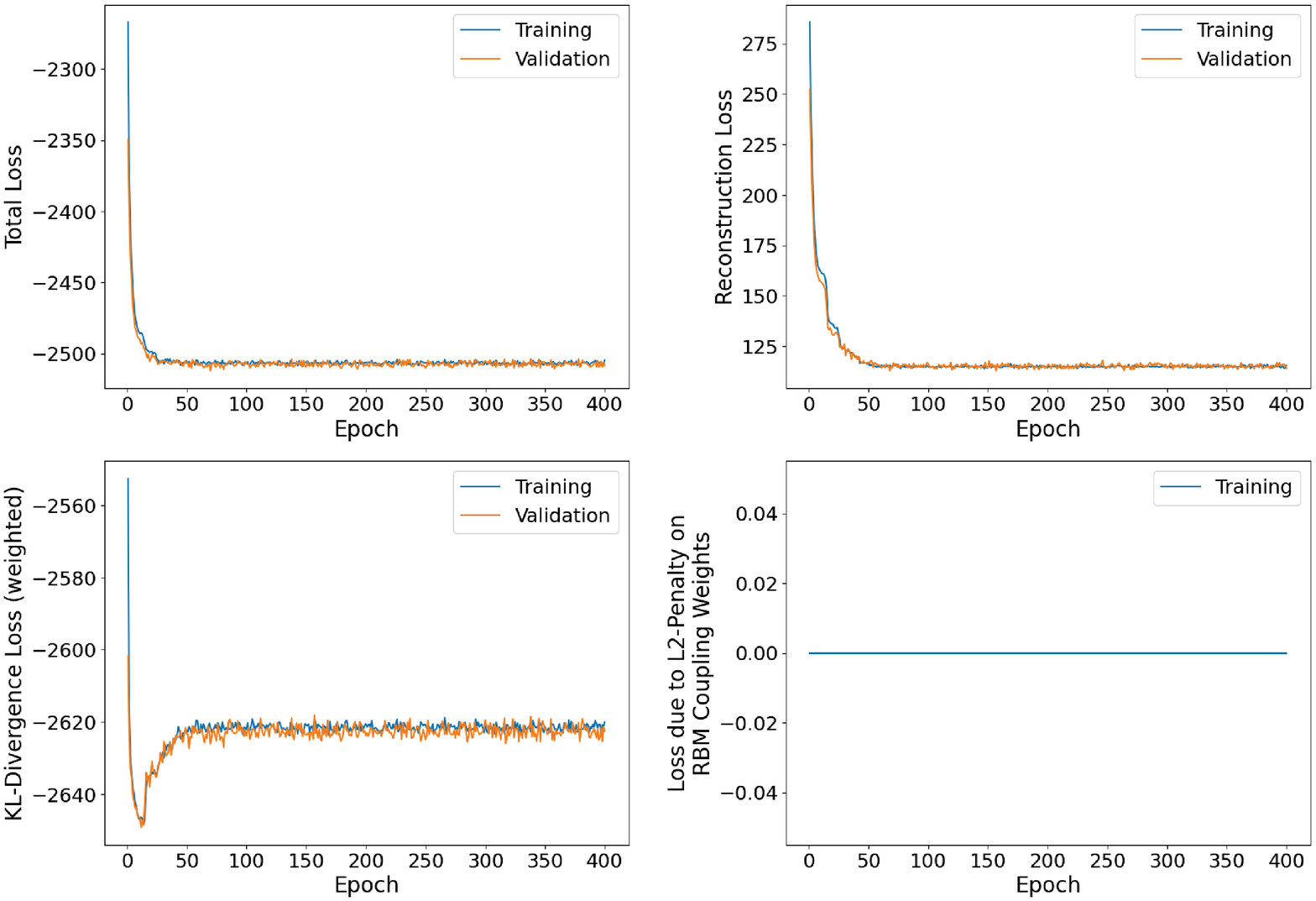}
    \let\nobreakspace\relax
    \caption{Typical total loss (negative of $\beta$-ELBO) as well as its two components reconstruction loss and KL-divergence loss during training and validation for 400 epochs. The KL-divergence loss (weighted with $\beta$) is negative because the log probability of the RBM prior, $\mathrm{log} \; p_{\bm{\theta}}(\mathbf{z})$, is not normalized. The loss due to the $\mathrm{L}_2$ penalty on the RBM's weight matrix \textbf{W} is zero because this feature, which did not produce any performance gain, was not used during the training of the VAE/RBM model.}
    \addtocounter{figure}{-1}
    \phantomcaption
    \label{fig:1s}
\end{figure}

\begin{figure}[!htb]
    \captionsetup{singlelinecheck = true, justification=justified, font=footnotesize, labelsep=period, width=1\textwidth}
    \centering
    \includegraphics[width=1\textwidth]{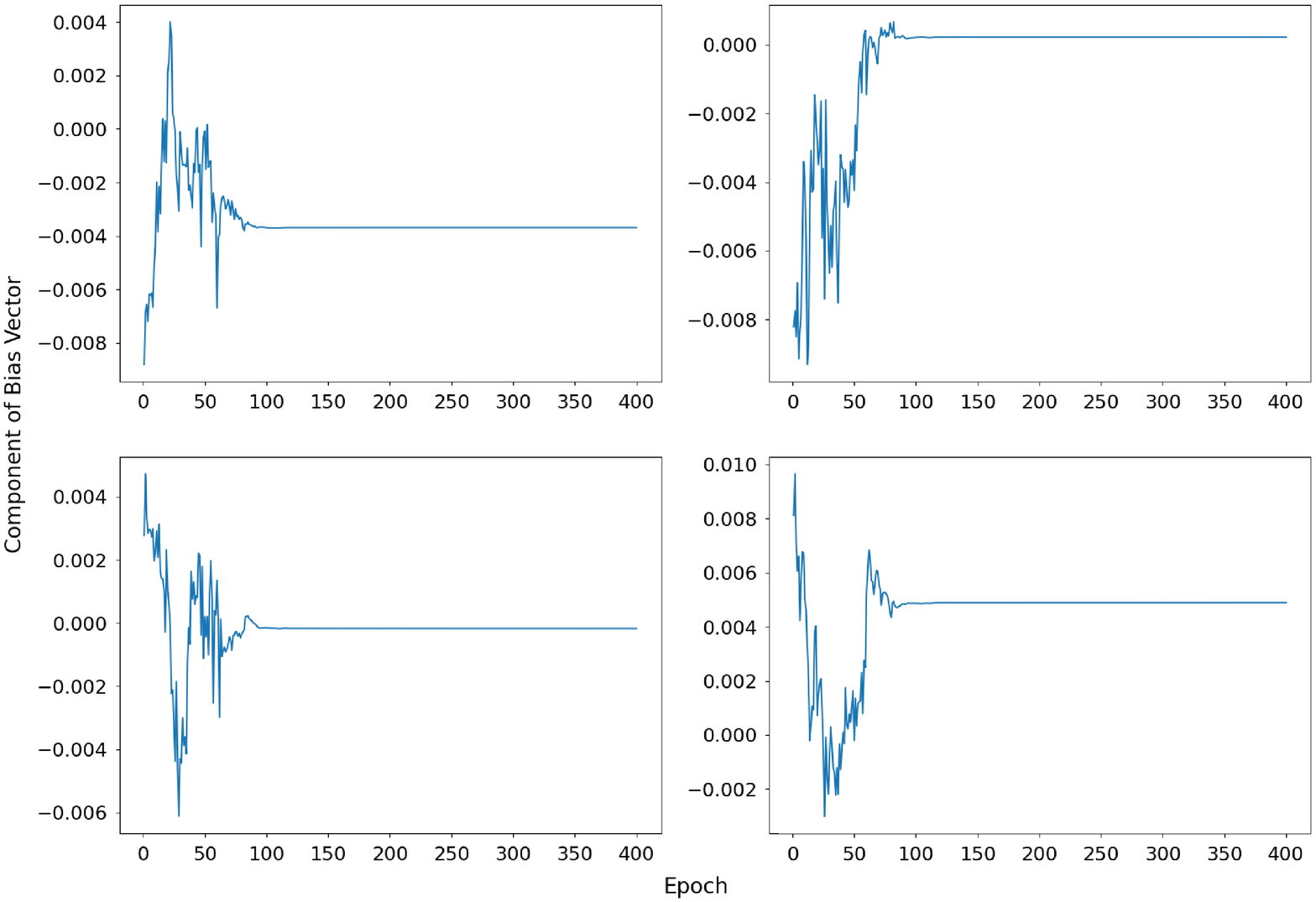}
    \let\nobreakspace\relax
    \caption{Evolution of the first four components of the RBM's bias vector \textbf{a} during model training, suggesting a dynamic exploration of the energy landscape of the system's configuration space.}
    \addtocounter{figure}{-1}
    \phantomcaption
    \label{fig:2s}
\end{figure}

\begin{figure}[!htb]
    \captionsetup{singlelinecheck = true, justification=justified, font=footnotesize, labelsep=period, width=1\textwidth}
    \centering
    \includegraphics[width=1\textwidth]{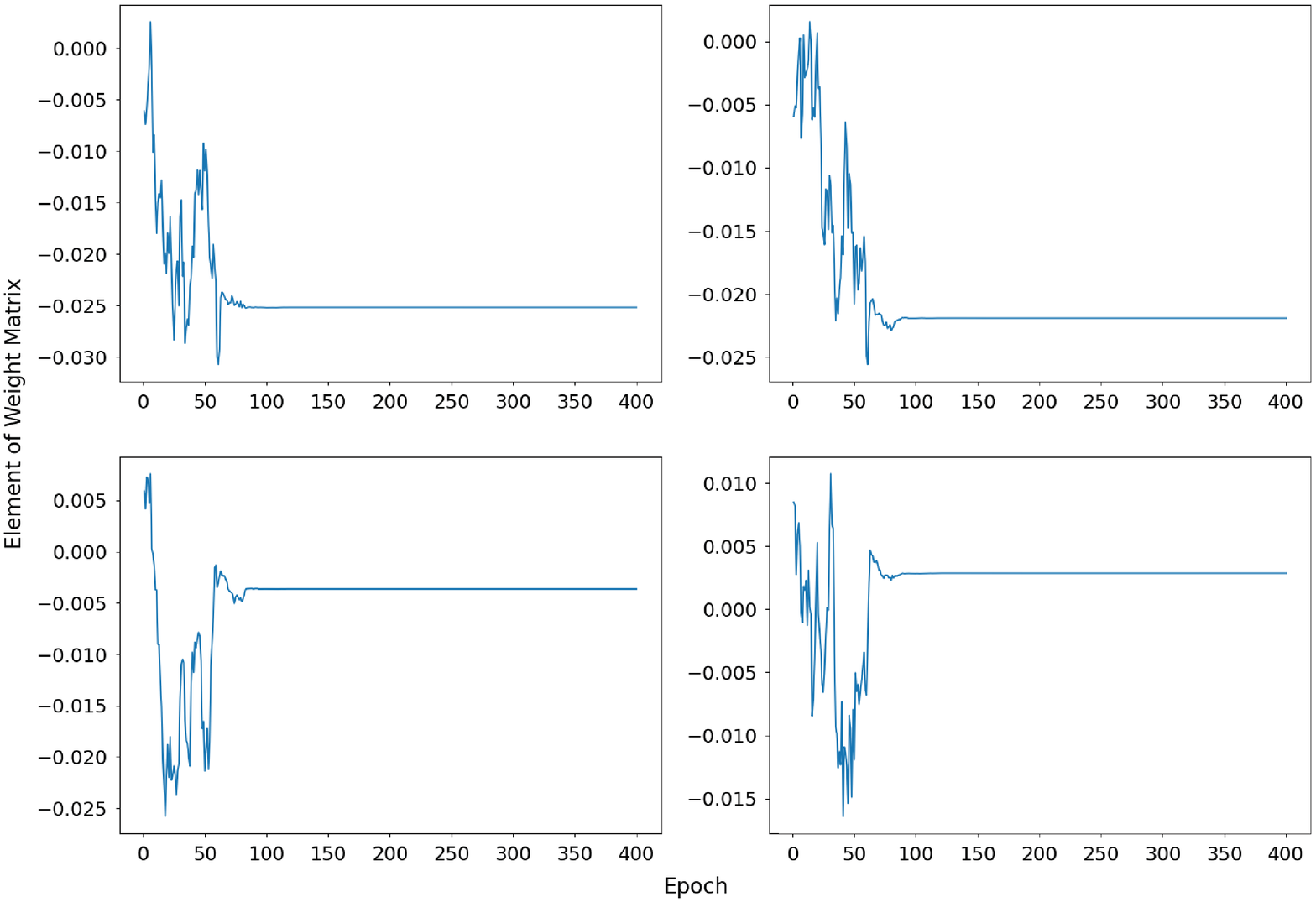}
    \let\nobreakspace\relax
    \caption{Evolution of the first four elements of the first column of the RBM's weight matrix \textbf{W} during model training, suggesting a dynamic exploration of the energy landscape of the system's configuration space.}
    \addtocounter{figure}{-1}
    \phantomcaption
    \label{fig:3s}
\end{figure}

\begin{figure}[!htb]
    \captionsetup{singlelinecheck = true, justification=justified, font=footnotesize, labelsep=period, width=1\textwidth}
    \centering
    \includegraphics[width=1\textwidth]{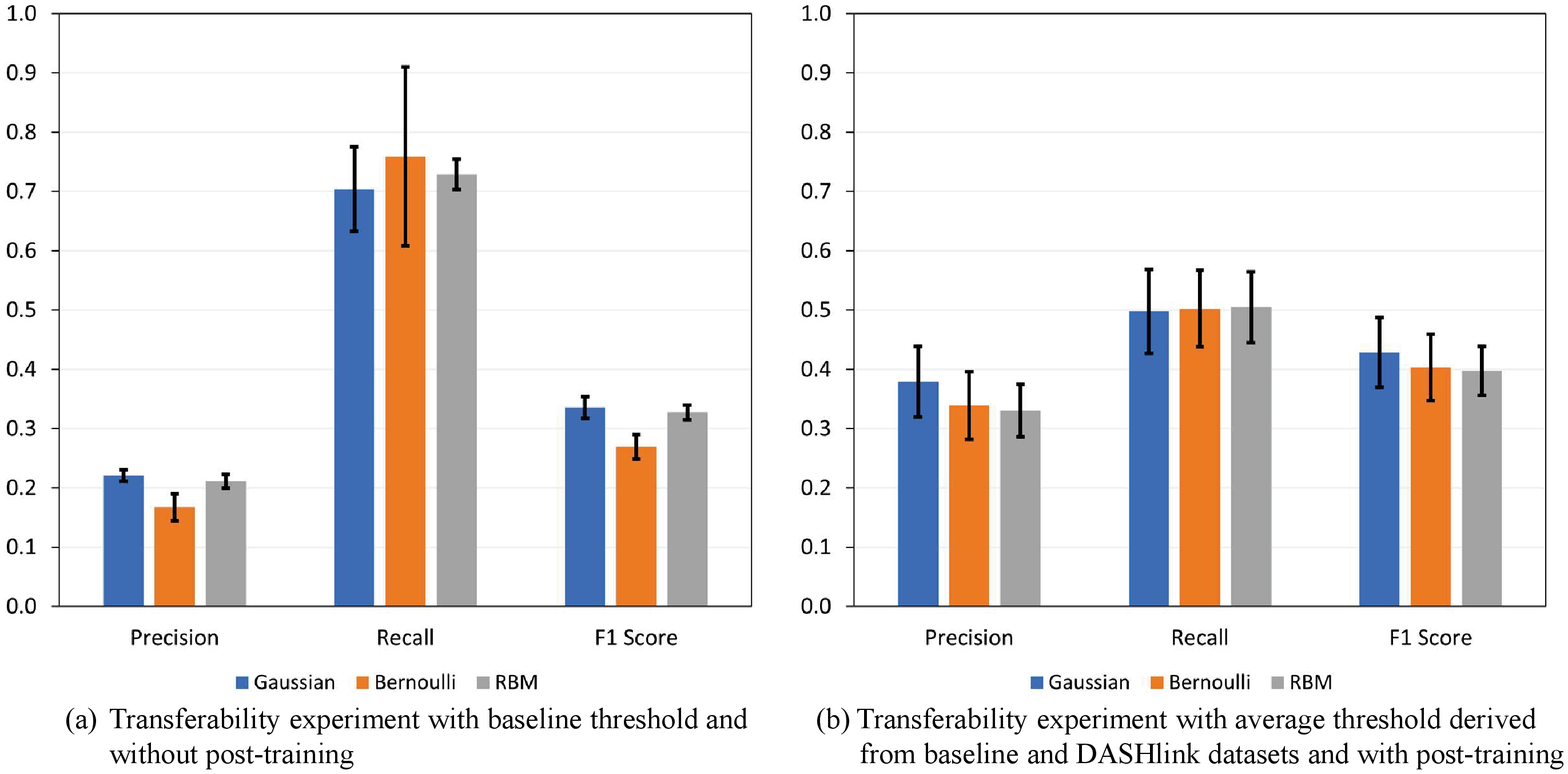}
    \let\nobreakspace\relax
    \caption{Performance of VAE models with Gaussian, Bernoulli, and RBM priors in the study examining the ability of the models tuned and trained on the baseline data to transfer to the DASHlink takeoff data. Error bars indicate standard deviations of performance metrics among 16 independently trained models. Performance was assessed based on (a) a threshold derived from the baseline training set, without post-training, or (b) the average value of the thresholds derived from the baseline and DASHlink/takeoff datasets, with post-training on the new dataset for 300 epochs with model weights initialized by training to convergence on the baseline dataset. The figure demonstrates an imbalance between precision and recall (recall $\gg$ precision) in the experiment without post-training and, in the post-training condition, similar findings as in the post-training experiment with a threshold based on the DASHlink/takeoff training set.}
    \addtocounter{figure}{-1}
    \phantomcaption
    \label{fig:4s}
\end{figure}

\begin{figure}[!htb]
    \captionsetup{singlelinecheck = true, justification=justified, font=footnotesize, labelsep=period, width=1\textwidth}
    \centering
    \includegraphics[width=1\textwidth]{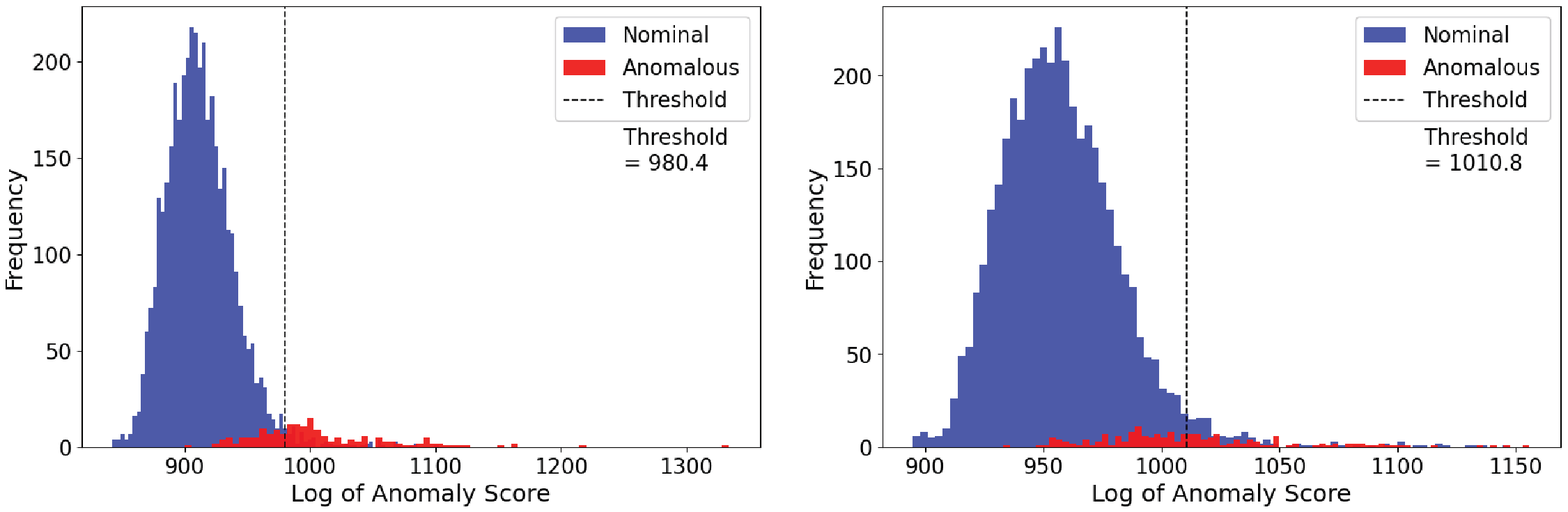}
    \let\nobreakspace\relax
    \caption{Histograms of average anomaly scores for two training modes of the RBM model trained and tested on the dataset with late-deployment-of-flaps anomaly during approach to landing. Data points to the right of the dashed threshold line are categorized as anomalies. The mode with a threshold of $\sim$980 demonstrates a good separation of nominal and anomalous data, whereas the mode with a threshold of $\sim$1011 exhibits a less good separation, with a relatively high number of anomalous data classified as nominal (false negatives).}
    \addtocounter{figure}{-1}
    \phantomcaption
    \label{fig:5s}
\end{figure}

\bibliography{references_supplement} \label{bibliographySuppl}